\documentclass[journal]{IEEEtran}

\usepackage{algorithm}
\usepackage{algorithmicx}
\usepackage{graphicx}
\usepackage{epstopdf}
\usepackage{subfigure}
\usepackage{amsmath}
\usepackage{multirow}
\usepackage{multicol}
\usepackage{url}
\usepackage{tabularx}
\usepackage{color}
\usepackage{amsmath}

\usepackage[numbers,sort]{natbib}

\ifCLASSINFOpdf
\else
\fi

\begin{document}
\title{Continuous Dropout}

\author{Xu~Shen,
        Xinmei~Tian,
        Tongliang~Liu,
        Fang~Xu,
        and~Dacheng~Tao % <-this % stops a space
\thanks{X. Shen, X. Tian are with the department of Electronic Engineering and Information Science,
University of Science and Technology of China (email: shenxu@mail.ustc.edu.cn;
xinmei@ustc.edu.cn).}% <-this %
\thanks{F. Xu is with the department of Biophysics and Neurobiology,
School of Life Science, University of Science and Technology of China
(email: xufan@mail.ustc.edu.cn).}
\thanks{T. Liu and D. Tao are with the Centre for Quantum Computation \& Intelligent
Systems and the Faculty of Engineering and Information Technology, University
of Technology, Sydney, 81 Broadway Street, Ultimo, NSW 2007, Australia (email:
tongliang.liu@uts.edu.au; dacheng.tao@uts.edu.au).}% <-this % stops a space
}

% \markboth{IEEE TRANSACTIONS ON NEURAL NETWORKS AND LEARNING SYSTEMS,~Vol.~XX, No.~XX, XXX~2016}%
% {Shell \MakeLowercase{\textit{et al.}}: Bare Demo of IEEEtran.cls for IEEE Journals}

\maketitle

\begin{abstract}
Dropout has been proven to be an effective algorithm for training robust deep
networks because of its ability to prevent overfitting by avoiding the
co-adaptation of feature detectors. Current explanations of dropout include bagging, naive
Bayes, regularization, and sex in evolution. According to the activation
patterns of neurons in the human brain, when faced with different situations, the
firing rates of neurons are random and continuous, not binary as current dropout
does. Inspired by this phenomenon, we extend the traditional binary dropout to
continuous dropout. On the one hand, continuous dropout is considerably closer to the
activation characteristics of neurons in the human brain than traditional
binary dropout. On the other hand, we demonstrate that continuous dropout has the
property of avoiding the co-adaptation of feature detectors, which suggests that we
can extract more independent feature detectors for model averaging in the test
stage. We introduce the proposed continuous dropout to a feedforward neural
network and comprehensively compare it with binary dropout, adaptive dropout,
and DropConnect on MNIST, CIFAR-10, SVHN, NORB, and ILSVRC-12. Thorough experiments demonstrate
that our method performs better in preventing the co-adaptation of feature
detectors and improves test performance. The code is available at:
\url{https://github.com/jasonustc/caffe-multigpu/tree/dropout}.
\end{abstract}


% \fi
\IEEEpeerreviewmaketitle

\section{Introduction}

\IEEEPARstart{D}{ropout} is an efficient algorithm introduced by Hinton \emph{et al} for
training robust neural networks \cite{hinton1} and  has been applied to many vision tasks \cite{tnn-15-scene, tnn-15-qa, AlexNet}. During the training stage,
hidden units of the neural networks are randomly omitted at a rate of 50\% \cite{hinton1}\cite{tnn:learningdeep}.
Thus, the presentation of each training sample can be viewed as providing
updates of parameters for a randomly chosen subnetwork. The weights of this
subnetwork are trained by back propagation \cite{spieg2}. Weights are shared
for the hidden units that are present among different subnetworks at each
iteration. During the test stage, predictions are made by the entire network, which
contains all the hidden units with their weights halved.

The motivation and intuition behind dropout is to prevent overfitting by
avoiding co-adaptations of the feature detectors \cite{hinton1}.
Deep network can achieve better representation than shallow networks,
but overfitting is a serious problem when training a large feedforward neural network on a small
training set \cite{hinton1}\cite{tnn:deepandshallow}. Randomly dropping the units from the neural
network can greatly reduce this overfitting problem. Encouraged by the success
of dropout, several related works have been presented, including fast
dropout \cite{wang10}, adaptive dropout \cite{ba13}, and DropConnect
\cite{wan14}. To accelerate dropout training, Wang and Manning suggested sampling
the output from an approximated distribution rather than sampling binary mask
variables for the inputs \cite{wang10}. In \cite{ba13}, Ba and Frey proposed
adaptively learning the dropout probability $p$ from the inputs and weights of
the network. Wan \emph{et al}. generalized dropout by randomly dropping the
weights rather than the units \cite{wan14}.

To interpret the success of dropout, several explanations from both theoretical and biological perspectives have been proposed. Based
on theoretical explanations, dropout is viewed as an extreme form of bagging
\cite{hinton1}, as a generalization of naive Bayes \cite{hinton1}, or as
adaptive regularization \cite{wager5}\cite{baldi6}, which is proven to be
a very useful approach for neural network training \cite{tnn:regularization}. From the biological
perspective, Hinton \emph{et al}. explain that there is an intriguing similarity
between dropout and the theory of the role of sex in evolution \cite{hinton1}.
However, no understanding from the perspective of the brain's neural network
$-$ the origin of deep neural networks $-$ has been proposed. In fact, by
analyzing the firing patterns of neural networks in the human brain
\cite{buzsaki7}\cite{fatt8}\cite{hippocampal_neuron},
we find that there is a strong analogy between
dropout and the firing pattern of brain neurons. That is, a small minority of
strong synapses and neurons provide a substantial portion of the activity in
all brain states and situations \cite{buzsaki7}. This phenomenon explains why
we need to randomly delete hidden units from the network and train different
subnetworks for different samples (situations). However, the remainder of the brain
is not silent. The remaining neuronal activity in any given time window is
supplied by very large numbers of weak synapses and cells. The amplitudes of
oscillations of neurons obey a random continuous pattern \cite{fatt8}\cite{hippocampal_neuron}. In other
words, the division between ``strong'' and ``weak'' neurons is not absolute. They obey a continuous $-$ rather than bimodal $-$ distribution \cite{fatt8}. Consequently, we should assign a continuous random mask to each neuron in the
dropout network for the divisions of ``strong'' and ``weak'' rather than use a
binary mask to choose ``activated'' and ``silent'' neurons.
%
%\begin{figure*}[t!]
%\centering
%    \subfigure[Distribution of amplitudes of individual units in nerve terminals\cite{fatt8}]{
%        \includegraphics[height=1.5in]{figs/1.eps}}
%    \hspace{6mm}
%    \subfigure[Variability of miniature excitatory postsynaptic current evoked in hippocampal neurons\cite{hippocampal_neuron}]{
%        \includegraphics[height=1.5in]{figs/hippocampal.eps}}
%\caption{Distribution of amplitudes of an individual neuron in neural networks.}
%\label{fig:bio.drop}
%%\framebox[4.0in]{$\;$}
%%\includegraphics[width=0.3\textwidth]{figs/hippocampal.eps}
%%\vspace{-6mm}
%\end{figure*}

Inspired by this phenomenon, we propose a continuous dropout algorithm in this
paper, \emph{i.e.}, the dropout variables are subject to a continuous
distribution rather than the discrete (Bernoulli) distribution in
\cite{hinton1}. Specifically, in our continuous dropout, the units
in the network are randomly multiplied by continuous dropout masks sampled from
$\mu \sim U(0,1)$ or $g \sim \mathcal{N}(0.5,\sigma^2)$, termed uniform dropout or Gaussian dropout, respectively.
Although multiplicative Gaussian noise has been mentioned in
\cite{JMLR:dropout}, no theoretical analysis or generalized continuous
dropout form is presented. We investigate two specific continuous
distributions, \emph{i.e.}, uniform and Gaussian, which are commonly used and
also are similar to the process of neuron activation in the brain. We conduct
extensive theoretical analyses, including both static and dynamic
property analyses of
our continuous dropout, and demonstrate that continuous dropout prevents the
co-adaptation of feature detectors in deep neural networks. In the static analysis,
we find that continuous dropout achieves a good balance between the diversity and
independence of subnetworks. In the dynamic analysis, we find that continuous
dropout training is equivalent to a regularization of covariance between
weights, inputs, and hidden units, which successfully prevents the co-adaptation of
feature detectors in deep neural networks.

We evaluate our continuous dropout through extensive experiments on several
datasets, including MNIST, CIFAR-$10$, SVHN, NORB, and ILSVRC-$12$.
We compare it with Bernoulli
dropout, adaptive dropout, and DropConnect. The experimental results demonstrate
that our continuous dropout performs better in preventing the co-adaptation of
feature detectors and improves test performance.

\section{Continuous Dropout}
%Our continuous dropout is inspired by the activation pattern of brain neurons.
%Figure \ref{fig:bio.drop} from \cite{fatt8}\cite{hippocampal_neuron} shows
%the distribution of the activation amplitudes
%of a single unit in neural networks. This figure indicates that the amplitudes are
%continuous and are generally found to be scattered in an approximately normal
%manner around a simple mean value. Because the division of ``strong'' and ``weak''
%neurons for different situations in the intact brain is not absolute
%\cite{buzsaki7}, using the continuous dropout variable as weights should be
%considerably closer to the way the brain neuron works than using discrete/binary
%dropout in \cite{hinton1}. Specifically, in our continuous dropout, the units
%in the network are randomly multiplied by continuous dropout masks sampled from
% $\mu \sim U(0,1)$ or $g \sim \mathcal{N}(0.5,\sigma^2)$, termed uniform dropout or Gaussian dropout, respectively.

In \cite{hinton1}, Hinton \emph{et al}. interpret dropout from the biological
perspective, \emph{i.e.}, it has an intriguing similarity to the theory of the
role of sex in evolution \cite{lehmann9}. Sexual reproduction involves taking
half the genes of each parent and combining them to produce
offspring. This corresponds to the result where dropout training works the best
when $p=0.5$; more extreme probabilities produce worse results \cite{hinton1}.
The criteria for natural selection may not be individual fitness but rather
the mixability of genes to combine \cite{hinton1}. The ability of genes to work
well with another random set of genes makes them more robust. The mixability
theory described in \cite{Livnat25} is that sex breaks up sets of co-adapted
genes, and this means that achieving a function by using a large set of
co-adapted genes is not nearly as robust as achieving the same function,
perhaps less than optimally, in multiple alternative ways, each of which only
uses a small number of co-adapted genes.

Following this train of thought, we can infer that randomly dropping units
tends to produce more multiple alternative networks, which is able to achieve
better performance. For example, when we use one hidden layer with $n$ units
for dropout training, \emph{i.e.}, the value of the dropout variable
is randomly set to $0$
or $1$, $2^n$ alternative networks will be produced during training
and will make up the entire network for testing. From this perspective, it is
more reasonable to take the continuous dropout distribution into account because, for
continuous dropout variables, a hidden layer with $n$ units can produce an
infinite number of multiple alternative networks, which are expected to work
better than the Bernoulli dropout proposed in \cite{hinton1}. The experimental
results in Section IV demonstrate the superiority of continuous dropout over
Bernoulli dropout.

\section{Co-adaptation Regularization in Continuous Dropout}

In this section, we derive the static and dynamic properties of our continuous
dropout. Static properties refer to the properties of the network with a fixed
set of weights, that is, given an input, how dropout affects the output of the
network. Dynamic properties refer to the properties of updating of the weights
for the network, \emph{i.e.}, how continuous dropout changes the learning
process of the network \cite{baldi6}. Because Bernoulli dropout with $p=0.5$
achieves the best performance in most situations
\cite{hinton1}\cite{srivastava16}, we set $p=0.5$ for Bernoulli dropout. For
our continuous dropout, we apply $\mu \sim U(0,1)$ and $g \sim \mathcal{N}(0.5,
\sigma^2)$ for uniform dropout and Gaussian dropout to ensure that all three
dropout algorithms have the same expected output (0.5).

\subsection{Static Properties of Continuous Dropout}
In this section, we focus on the static properties of continuous dropout,\emph{
i.e.}, properties of dropout for a fixed set of weights. We start from the
single layer of linear units, and then we extend it to multiple layers of linear and
nonlinear units.

\subsubsection{Continuous dropout for a single layer of linear units}

We consider a single fully connected linear layer with input $\mathbf{I} = [I_1,
I_2,..., I_n]^T$, weighting matrix $W = [w_{ij}]_{k\times n}$, and output $\mathbf{S} =
[S_1, S_2,..., S_k]^T$. The $i$th output $S_i = \sum \nolimits_{j=1}^n
w_{ij}I_j$. In Bernoulli dropout, each input unit  $I_j$ is kept with
probability $p \sim Bernoulli(0.5)$. The $i$th output and its expectation
are
$$S_i^B = \sum \limits_{j=1}^n w_{ij}I_jp_j ~~\text{and} ~~
\textbf{E}[S_i^B] = \frac{1}{2} \sum \limits_{j=1}^nw_{ij}I_j.$$
In our uniform dropout, $I_j$ is kept with probability $u \sim U(0,1)$. The
output becomes
$$S_i^U = \sum \limits_{j=1}^nw_{ij}I_ju_j ~~\text{and} ~~
\textbf{E}[S_i^U] = \frac{1}{2} \sum \limits_{j=1}^nw_{ij}I_j.$$
When Gaussian dropout is applied, $I_j$ is kept with probability $g \sim \mathcal{N}(0.5,
\sigma^2)$,
$$S_i^G = \sum \limits_{j=1}^nw_{ij}I_jg_j ~~\text{and} ~~
\textbf{E}[S_i^G] = \frac{1}{2} \sum \limits_{j=1}^nw_{ij}I_j.$$
Therefore, the three dropout methods achieve the same expected output.

Because dropout is applied to the input units independently, the variance and covariance of the output units are:
\begin{equation}
\begin{aligned}
&Var(S_i^U)=\sum \limits_{j=1}^{n}w_{ij}^2I_j^2Var(u_j)=\sum \limits_{j=1}^nw_{ij}^2I_j^2\frac{1}{12}, \nonumber \\
&Cov(S_i^U,S_l^U)=\sum \limits_{j=1}^nw_{ij}w_{lj}I_j^2\frac{1}{12}, \nonumber \\
&Var(S_i^G)=\sum \limits_{j=1}^n w_{ij}^2I_{j}^2Var(g_i) = \sum \limits_{j=1}^n w_{ij}^2I_j^2\sigma^2, \nonumber \\
&Cov(S_i^G,S_l^G)=\sum \limits_{j=1}^n w_{ij}w_{lj}I_j^2\sigma^2, \nonumber \\
&Var(S_i^B) = \sum \limits_{j=1}^n w_{ij}^2I_j^2p_jq_j = \sum \limits_{j=1}^{n}w_{ij}^2I_j^2\frac{1}{4}, \nonumber \\
&Cov(S_i^B,S_l^B)=\sum \limits_{j=1}^{n}w_{ij}w_{lj}I_j^2p_jq_j=\sum \limits_{j=1}^{n}w_{ij}w_{lj}I_j^2\frac{1}{4}. \nonumber
\end{aligned}
\end{equation}

The aim of dropout is to avoid the co-adaptation of feature detectors,
    reflected by the covariance between output units. Generally, networks with
    lower covariance between feature detectors tend to generate more
    independent subnetworks and therefore tend to work better during the test
    stage. Comparing the covariance of the output units of the three dropout
    algorithms, we can see that uniform dropout has a lower covariance than
    Bernoulli dropout. The covariance of Gaussian dropout is controlled by
    the parameter $\sigma^2$. Through extensive experiments, we find that Gaussian
    dropout with $\sigma^2 \in [\frac{1}{5},\frac{1}{3}]$ works the best among the
three dropout algorithms. This phenomenon implies that there is a balance
between the diversity of subnetworks (larger variance of the output of hidden
units) and their independence (lower covariance between units in the same
layer). Bernoulli dropout achieves the highest variance but its covariance is
also the highest. In contrast, uniform dropout achieves the lowest covariance, but its
variance is also the lowest. Gaussian dropout with a suitable $\sigma^2$
achieves the best balance between variance and covariance, ensuring a good
generalization capability.

\subsubsection{Continuous dropout approximation for non-linear unit}

\label{headings}
For the non-linear unit, we consider the case that the output of a single unit with total linear input $S$ is given by the logistic sigmoidal function
\begin{equation}
O=sigmoid(S) = \frac{1}{1+ce^{-\lambda S}}.
\end{equation}

For uniform dropout, $S = \sum \limits_{i=1}^nw_iI_iu_i, u_i \sim U(0,1)$. We have $S= \sum \limits_{i=1}^nU_i, U_i \sim U(0, w_iI_i),\textbf{E}[U_i]= \frac{1}{2}w_iI_i$ and $ Var(U_i)= \frac{1}{12}w_i^2I_i^2$. Because $U_i \leq \max \limits_i (w_iI_i)$, $s_n^2 = \sum \limits_{i=1}^nVar(U_i)\rightarrow \infty $. According to Corollary 2.7.1 of Lyapunov's central limit theorem \cite{lehmann9}, $S$ tends to a normal distribution as $n \rightarrow \infty$. It yields that

\begin{equation}
\begin{aligned}
&S \sim \mathcal{N}(\mu_U, \sigma_U^2), \\
&\mu_U = \sum \limits_{i=1}^n\textbf{E}[U_i] = \sum \limits_{i=1}^n \frac{1}{2}w_iI_i, \\
&\sigma_U^2 = \sum \limits_{i=1}^n Var(U_i) = \sum \limits_{i=1}^n \frac{1}{12}w_i^2I_i^2
\end{aligned}
\end{equation}

For Gaussian dropout, $S = \sum \limits_{i=1}^n w_i I_i g_i, g_i \sim \mathcal{N}(\mu, \sigma^2)$ and $g_i$ i.i.d. We can easily infer that $S \sim \mathcal{N}(\mu_S, \sigma_S^2)$, where $\mu_S = \sum \limits_{i=1}^n w_i I_i \mu, \sigma_S^2 = \sum \limits_{i=1}^n w_i^2 I_i^2 \sigma^2$.

jhus, for both uniform dropout and Gaussian dropout, $S$ is subject to a normal distribution. In the following sections, we only derive the statistical property of Gaussian dropout because it is the same for uniform dropout.

The expected output is
\begin{equation}
\begin{aligned}
\textbf{E}(O) &= \textbf{E}[sigmoid(S)] \\
&= \int_{-\infty}^{\infty}sigmoid(x)\mathcal{N}(x|\mu_S,\sigma_S^2)\,dx \\
&\approx sigmoid(\frac{\mu_S}{\sqrt{1 + \pi \sigma_S^2/8}}).
\end{aligned}
\end{equation}
This means that for Gaussian dropout $g_i \sim \mathcal{N}(\mu, \sigma^2)$, we have the recursion

\begin{equation}
\begin{aligned}
\textbf{E}[S_i^h] = \sum \limits_{l < h} \sum \limits_{j}w_{ij}^{hl}\textbf{E}[g_j^l]\textbf{E}[O_j^l] \quad and \\
\quad  \textbf{E}[O_i^h] \approx sigmoid(\frac{\textbf{E}[S_i^h]}{1+ \pi Var(S_i^h)/8}).
\end{aligned}
\end{equation}

While for Bernoulli dropout \cite{baldi6}:
\begin{equation}
\begin{aligned}
\textbf{E}[S_i^h] &= \sum \limits_{l < h} \sum \limits_{j} w_{ij}^{hl}\textbf{ E}[\delta_j^l]\textbf{E}[O_j^l] \\
\textbf{E}[O_i^h] &\approx sigmoid(\textbf{E}[S_i^h]).
\end{aligned}
\end{equation}
In Bernoulli dropout, the expected output is only the propagation of deterministic variables among the entire network, whereas our continuous dropout has a regularization term of $\sqrt{1+\pi Var(S_i^h)/8}= \sqrt{1+ \pi (\sum \limits_{i=1}^nw_i^2I_i^2\sigma^2)/8}$. Thus, continuous dropout can regularize complex weights and inputs during forward propagation.

\subsection{Dynamic Properties of Continuous Dropout}

In this section, we will investigate the dynamic properties of continuous
dropout related to the training procedure and the update of the weights. We
also start from the simple case of a single linear unit, and then we discuss the
non-linear case. As proven in last section, in uniform dropout, the $S$ tends
to a normal distribution as $n \rightarrow \infty$. Therefore, we analyze the
dynamic properties of Gaussian dropout only.

\subsubsection{Continuous dropout gradient and adaptive
  regularization -- single linear unit}

In the case of a single linear unit trained with dropout with an input $I$, an
output $O=S$, and a target $t$, the error is typically quadratic of the form
$E_D = \frac{1}{2}(t-O)^2$, where $O = S = \sum \nolimits_{i}w_ip_iI_i$.  In
the linear case, the ensemble network is identical to the deterministic network
obtained by scaling the connections using the dropout probabilities. For a
single output $O$, the ensemble error of all possible subnetworks $E_{ENS}$ is
defined by:
$$ E_{ENS} =\frac{1}{2}(t-O_{ENS})^2=\frac{1}{2}(t-\sum_{i=1}^n \mu w_iI_i)^2.$$
%\begin{equation}
%E_{ENS} = \frac{1}{2}\textbf{E}[(t-O)^2].
%\end{equation}
The gradients of the ensemble error can be computed by:
\begin{equation}
    \frac{\partial E_{ENS}}{\partial w_i} = -(t-O_{ENS})\mu I_i .
\end{equation}

%\begin{equation}
%E_D  = \frac{1}{2}(t - \sum \nolimits_{i=1}^n g_iw_iI_i)^2.
%\end{equation}
For Gaussian dropout, $E_D  = \frac{1}{2}(t - \sum \nolimits_{i=1}^n g_iw_iI_i)^2$.
Here, $g \sim \mathcal{N}(\mu, \sigma^2)$ is the random variable with a Gaussian distribution. Hence, $E_D$ is a random variable, while $E_{ENS}$ is a deterministic function.
%Since $\mu_O = \textbf{E}(O) = \sum \nolimits_i w_i\mu I_i, \sigma_O^2 = Var(O) = \sum \nolimits_iw_i^2I_i^2\sigma^2$, it yields that:
%\begin{equation}
%\begin{aligned}
%E_{ENS} &= \int_{O =- \infty}^{\infty} \frac{1}{2}(t-O)^2N(\mu_O, \sigma_O^2)dO \\
%&= \int_{O =- \infty}^{\infty} \frac{1}{2}(t^2 - 2tO + O^2)\frac{1}{2 \pi \sigma_O^2}e^{-\frac{1}{2 \sigma_O^2}(O-\mu_O)^2}dO \\
%&= \frac{1}{2}t^2 - t\int_{O =- \infty}^{\infty} \frac{1}{2\pi \sigma_O^2} e^{- \frac{1}{2\sigma_O^2}(O-\mu_O)^2}dO+\frac{1}{2}\int_{O =- \infty}^{\infty} O^2 \frac{1}{2\pi \sigma_O^2} e^{- \frac{1}{2\sigma_O^2}(O-\mu_O)^2}dO \\
%&= \frac{1}{2}t^2 - t\mu_O+ \frac{1}{2}\textbf{E}[O^2] \\
%&= \frac{1}{2}t^2 -t\mu_O + \frac{1}{2}(Var(O)+(\textbf{E}[O])^2) \\
%&= \frac{1}{2}t^2 - t\mu_O + \frac{1}{2} \sigma_O^2 + \frac{1}{2} \mu_O^2 \\
%&= \frac{1}{2}t^2 - t\sum \nolimits_i w_i\mu I_i + \frac{1}{2}\sum \nolimits_i w_i^2I_i^2\sigma^2 + \frac{1}{2}(\sum \nolimits_i w_i\mu I_i)^2
%\end{aligned}
%\end{equation}

%Gradients of the ensemble error can be computed by:
%\begin{equation}
%\frac{\partial E_{ENS}}{\partial w_i} = -t\mu_iI_i + w_iI_i^2\sigma_i^2 + \frac{1}{2} \sum \nolimits_{j\neq i} w_j\mu^2I_iI_j + w_i\mu^2I_i^2.
%\end{equation}

For dropout error, the learning gradients are of the form
$$\frac{\partial E_D}{\partial w_i} = \frac{\partial E_D}{\partial O} \frac{\partial O}{\partial w_i} = -(t-O)\frac{\partial O}{\partial w_i};$$
therefore,
\begin{equation}
\begin{aligned}
\frac{\partial E_D}{\partial w_i} &= -(t - O_D)g_iI_i \\
&= -tg_iI_i + w_ig_i^2I_i^2 + \sum \limits_{j \neq i} w_jg_ig_jI_iI_j,
\end{aligned}
\end{equation}
and
\begin{equation}
\begin{aligned}
\textbf{E}[\frac{\partial E_D}{\partial w_i}] &= -t\mu I_i + w_i(\mu^2 + \sigma^2)I_i^2 + \sum \limits_{j \neq i} w_j\mu^2I_iI_j \\
&= \frac{\partial E_{ENS}}{\partial w_i} + w_iI_i^2\sigma^2.
\end{aligned}
\end{equation}
Remarkably, the relationship between the expectation of ensemble error and dropout error is:
\begin{equation}
E_D = E_{ENS} + \frac{1}{2} \sum \limits_{i=1}^n w_i^2I_i^2\sigma^2.
\end{equation}

In Bernoulli dropout \cite{baldi6}, this relationship is:
\begin{equation}
E_D = E_{ENS} + \frac{1}{2}\sum \limits_{i=1}^nw_i^2I_i^2Var(p_i).
\end{equation}
Generally, the regularization term is weight decay based on the square of
    the weights, and it ensures that the weights do not become too large to
    overfit the training data. Bernoulli dropout extends this regularization
    term by incorporating the square of the input terms and the variance of the
    dropout variables; however, both the expected output and the weight of
    regularization term are determined by the dropout probability ($p$), i.e., there
    is no freedom for adjusting the model complexity to reduce overfitting.
    In contrast, in Gaussian dropout, we have an extra degree of freedom of $\sigma^2$
to achieve the balance between network output and model complexity.

\subsubsection{Continuous dropout gradient and adaptive regularization -- single sigmoidal unit}

\label{others}
In Gaussian dropout, for a single sigmoidal unit,
$$O = sigmoid(S)= \frac{1}{1+ ce^{-\lambda S}},$$
where $S=\sum \nolimits_i w_i g_i I_i$ and  $S_{ENS}=\textbf{E}[S]=\sum \nolimits_i w_i\mu I_i$ with $\sigma_S^2 = \sum \nolimits_i w_i^2I_i^2\sigma^2,\mu_S=\sum \nolimits_i w_i \mu I_i$. Commonly, we use relative entropy error
\begin{equation}
    E_D = -(tlogO + (1-t)log(1-O)).
\end{equation}
By the chain rule $\frac{\partial E_D}{\partial w_i} = \frac{\partial E_D}{\partial O}\frac{\partial O}{\partial S}\frac{\partial S}{\partial w_i}$, we obtain $$\frac{\partial E_D}{\partial w_i} = -\lambda(t - O)\frac{\partial S}{\partial w_i}.$$
%\begin{equation}
%\frac{\partial E_D}{\partial w_i} = -\lambda(t - O)\frac{\partial S}{\partial w_i}.
%\end{equation}
For the ensemble network,
$$ \frac{\partial E_{ENS}}{\partial w_i} = -\lambda (t- O_{ENS})\frac{\partial S_{ENS}}{\partial w_i}.$$
%\begin{equation}
%\frac{\partial E_{ENS}}{\partial w_i} = -\lambda (t- O_{ENS})\frac{\partial S_{ENS}}{\partial w_i}.
%\end{equation}
We have
\begin{equation}
\begin{aligned}
O_{ENS} &= \textbf{E}[sigmoid(S)] \\
&= \int_{- \infty}^{\infty} \frac{e^S}{1+e^S}e^{-\frac{S-\mu_S^2}{2\sigma_S^2}}ds \\
&\approx sigmoid(\frac{\mu_S}{\sqrt{1+ \pi \sigma_S^2/8}}).
\end{aligned}
\end{equation}
Therefore,
\begin{equation}
\frac{\partial E_{ENS}}{\partial w_i} = -\lambda (t - sigmoid(\frac{\mu_S}{\sqrt{1+ \pi \sigma_S^2/8}}))\mu I_i.
\end{equation}
For the dropout network,
\begin{equation}
\frac{\partial E_D}{\partial w_i} = -\lambda(t - O)g_iI_i = -\lambda(t - sigmoid(\sum \nolimits_jw_jg_jI_j))g_iI_i.
\end{equation}
Here, $g_i$ are the random variables with Gaussian distributions; thus, $O_D = sigmoid(\sum \limits_j w_jg_jI_j)$ and $g_i$ are both random variables. It yields that
\begin{equation}
\textbf{E}[\frac{\partial E_D}{\partial w_i}] = \textbf{E}[-\lambda(t- sigmoid(\sum \nolimits_j w_jg_jI_j|g_j=\mu))\mu I_i],
\end{equation}
where $O_D^{'} = sigmoid(\sum \nolimits_j w_jg_jI_j|g_j = \mu), \textbf{E}[O_D^{'}]=\mu_S^{'} = \mu_S = \sum \nolimits_i w_i \mu I_i$, and $Var(O_D^{'})=\sum \nolimits_{j \neq i} w_j^2 I_j^2 \sigma^2$.
\begin{equation}
\textbf{E}[O_D^{'}] \approx sigmoid(\frac{\mu_S^{'}}{\sqrt{1+ \pi {\sigma_S^{'}}^2}/8}).
\end{equation}
The gradient of the dropout is:
\begin{equation*}
\begin{aligned}
&\textbf{E}[\frac{\partial E_D}{\partial w_i}] \approx \frac{\partial E_{ENS}}{\partial w_i} + \lambda \mu I_i (sigmoid(\frac{\mu_S}{\sqrt{1+ \pi \sigma_S^{'}/8}}) \\
& - sigmoid(\frac{\mu_S}{\sqrt{1 + \pi \sigma_S^2/8}})) \\
& \approx \frac{\partial E_{ENS}}{\partial w_i} + \lambda \mu I_i sigmoid^{'}(\frac{\mu_S}{\sqrt{1 + \pi \sigma_S^2/8}}) \\
&\times (\frac{\sqrt{1 + \pi \sigma_S^2/8}- \sqrt{1 + \pi {\sigma_S^{'}}^2/8}}{\sqrt{(1 + \pi {\sigma_S^{'}}^2/8)(1 + \pi \sigma_S^2/8)}}) \\
& \approx \frac{\partial E_{ENS}}{\partial w_i} + \lambda \mu I_i sigmoid^{'}(\frac{\mu_S}{\sqrt{1 + \pi \sigma_S^2/8}}) \\
& \times (\frac{\frac{\pi}{16}\mu_Sw_i^2I_i^2\sigma^2}{\sqrt{(1 + \pi {\sigma_S^{'}}^2/8)(1 + \pi \sigma_S^2/8)}})\\
& \approx \frac{\partial E_{ENS}}{\partial w_i} + \lambda \mu I_i sigmoid^{'}(\frac{\mu_S}{\sqrt{1 + \pi \sigma_S^2/8}}) \\
& \times (\frac{\frac{\pi}{16}\mu_Sw_i^2I_i^2\sigma^2}{1 + \pi \sigma_S^2/8}) \\
& = \frac{\partial E_{ENS}}{\partial w_i} + \lambda \mu I_i sigmoid^{'}(\frac{\mu_S}{\sqrt{1 + \pi \sigma_S^2/8}}) \\
&\times (\frac{\frac{\pi}{16}(\sum \nolimits_j w_j\mu I_j)(w_i^2I_i^2\sigma^2)}{1+ \frac{\pi}{8}\sum \limits_i w_i^2 I_i^2\sigma^2})\\
\end{aligned}
\end{equation*}

\begin{figure*}[htb!]
%\begin{figure*}[htb!f]
\begin{center}
\includegraphics[width=0.82\textwidth]{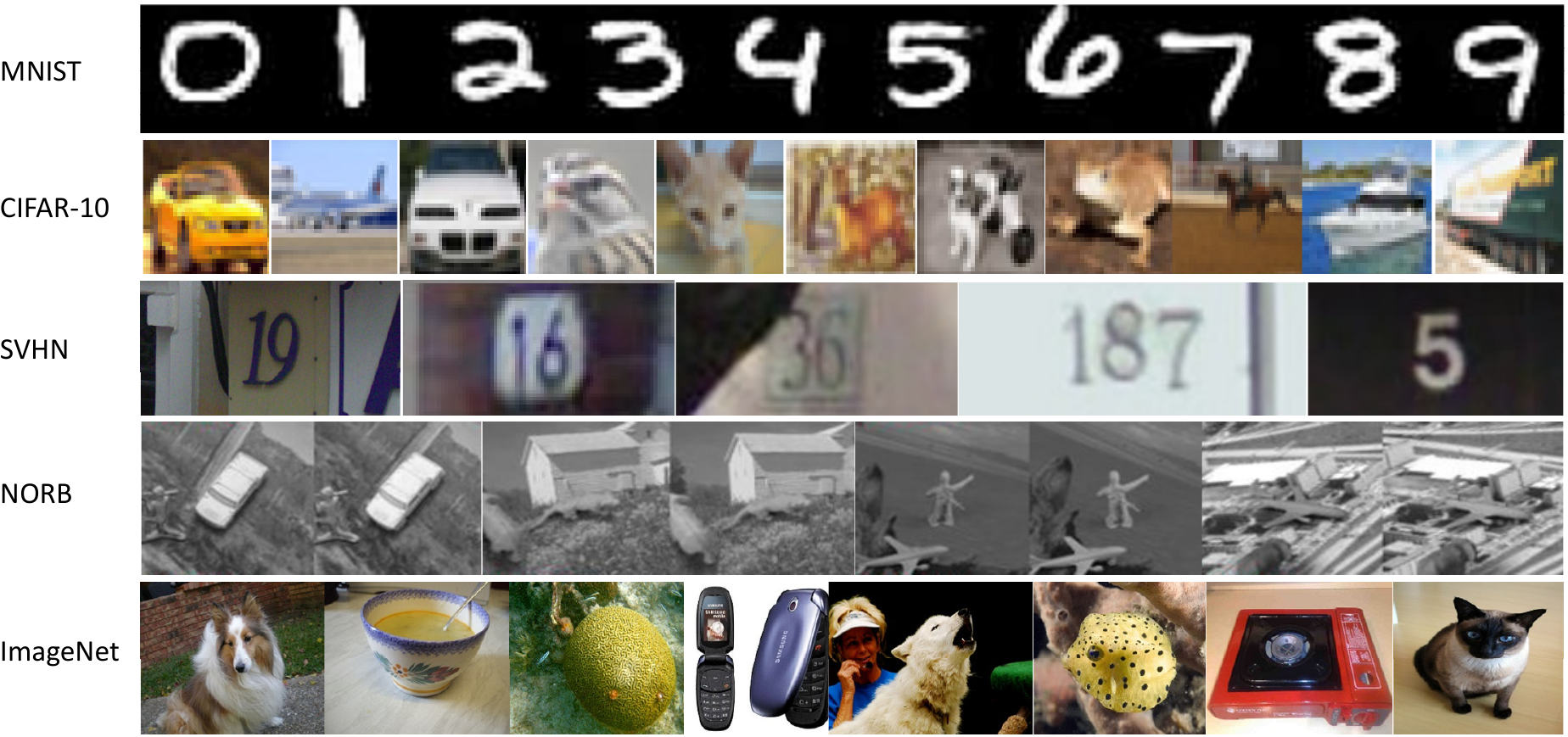}
\caption{ \scriptsize{Samples of benchmark datasets. MNIST and SVHN are digit classification
tasks. NORB, CIFAR-10, and ImageNet $2012$ are object recognition tasks.
All of them are formulated as classification problems, which is commonly evaluated by
classification accuracy (error).}}
\label{fig:dataset_sampls}
\end{center}
\end{figure*}

For approximation
\begin{equation}
\begin{aligned}
E_D &= E_{ENS} + \sum \limits_{i=1}^n \lambda w_i \mu_i I_i sigmoid^{'}(\frac{\mu_S}{\sqrt{1 + \pi \sigma_S^2/8}}) \\
&\times (\frac{\frac{\pi}{16}(\sum \limits_j w_j\mu I_j)(w_i^2I_i^2\sigma^2)}{1+ \frac{\pi}{8}\sum \limits_i w_i^2 I_i^2\sigma^2}) \\
& = E_{ENS} + \frac{1}{2} \lambda sigmoid^{'}(\frac{\mu_S}{\sqrt{1 + \pi \sigma_S^2/8}})\\
&\times \sum \limits_{i=1}^n \sum \limits_j w_iw_j\mu^2 I_iI_j(\frac{\pi}{8}w_i^2I_i^2 \sigma^2)/(1 + \frac{\pi}{8}\sum \limits_i w_i^2 I_i^2 \sigma^2)
\end{aligned}
\end{equation}
Note that for Bernoulli dropout \cite{baldi6}
\begin{equation}
E_D = E_{ENS} + \frac{1}{2}\lambda sigmoid^{'}(U)\sum \limits_{i=1}^n w_i^2 I_i^2 Var(p_i).
\end{equation}

Bernoulli dropout only provides the magnitude of the regularization term,
which is adaptively scaled by the square of the input terms, by the gain $\lambda$ of the
sigmoidal function, by the variance of the dropout variables, and by the
instantaneous derivative of the sigmoidal function; however, this
term only tends to achieve a simpler model and avoid overfitting. It has little
help in avoiding the co-adaptation of units (feature detectors) in the same
layer. In contrast, continuous dropout not only provides the regularization of squares of input units, weights,
and dropout variance individually $(\sum \nolimits_i w_i^2 I_i^2 \sigma^2)$, but
also regularizes the covariance between input units $(I_iI_j)$ and weights
$(w_i w_j)$. In other words, in Gaussian dropout, the regularization term penalizes the
covariance between weights, dropout variables, and input units; that is, it
prevents the co-adaptation of feature detectors in the neural network.
Therefore, through this co-adaptation regularization, Gaussian dropout can
indeed avoid co-adaptation and overfitting.

\section{Experiments}
We investigate the performance of our continuous dropout on MNIST
\cite{lecun18}, CIFAR-10 \cite{kriz24}, SVHN \cite{SVHN}, NORB \cite{NORB}, and ImageNet ILSVRC-2012
classification task \cite{ILSVRC15}. Samples and brief description of these datasets are presented
in Fig. \ref{fig:dataset_sampls}. We compare
continuous dropout with the original dropout proposed in \cite{hinton1}
(Bernoulli dropout), adaptive dropout \cite{ba13}, and DropConnect
\cite{wan14}. Fast Dropout \cite{wang10} is an approximation of Bernoulli
dropout that accelerates the sampling process. Its performance is similar to
that of Bernoulli dropout.
For evaluation metric, the classification error, which is defined as the ratio of misclassified samples to all samples, is applied ($0/1$ loss).
%Classification error ($0/1$ loss) is used as evaluation metric.
We use the publicly available THEANO library
\cite{bergstra12} to implement the feedforward neural networks that consist of
fully connected layers only, and the networks that consist of Convolutional
Neural Networks (CNNs) are
implemented based on Caffe \cite{jia17}.
In all experiments, the dropout rate
in Bernoulli dropout and DropConnect is set as $0.5$ because this is
the most commonly used configuration in dropout and performs the best.
All the other parameters are selected based on performance on the validation set.
To ensure that all three dropout algorithms achieve the same expected output, for uniform dropout, the variables
$u_i$ are subject to $U(0,1)$. In Gaussian dropout, $g_i \sim \mathcal{N}(0.5,\sigma^2)$, and
$\sigma^2$ is selected from $\{0.2, 0.3, 0.4\}$. For Adaptive dropout, $\alpha$ is selected
from $\{-1, 0, 1\}$ and $\beta$ is selected from $\{-0.5, 0, 0.5\}$.
To avoid divergence during propagation, we clip the Gaussian dropout variable
to be in $[0,1]$, yielding  $g_i=1$ if $g_i \geq 1$ and $g_i=0$ if $g_i
\leq 0$. 

To verify whether the performance gain is statistically significant,
we repeated all experiments $N$ times for all methods and reported the mean error and standard derivation.
Here, $N=30$ for datasets MNIST, CIFAR-10, SVHN, and NORB, and $N=10$ for dataset ImageNet ILSVRC-2012 because of the high computational cost in this dataset.
In each of the $N$ independent runs, we randomly initialized weights of the network and then applied different dropout algorithms to train this network. In other words, in the $i$th independent run ($i = 1,2,\cdots,N$), all dropout algorithms share the same weights initialization. In another independent run, the network was randomly initialized again, i.e., the network had different initialized weights in the $i$th run and the $j$th run ($i\neq j$). In this way, we obtained $N$ groups of results and then conducted paired t-test and paired Wilcoxon signed rank test between Gaussian dropout and all other baseline methods. Their p-values are reported.

%To verify whether the performance gain is statistically significant,
%we report mean and standard derivation of $5$ independent runs with different randomly
%initialized weights and conduct paired t-tests between Gaussian Dropout and all
%other baseline dropout methods.

\begin{table*}[t]
\label{table1}
\caption{Performance comparison on MNIST (mean error and standard derivation). No data augmentation is used. Architecture: $784-800-800-10$. Paired t-test and paired Wilcoxon signed rank test are conducted between Gaussian dropout and all other baseline methods. Their p-values are reported: p-value-T for t-test and p-value-W for Wilcoxon signed rank test.}
\begin{center}
\begin{tabular}{l|c|c}
\hline
\multirow{2}{*}{Method} & \multicolumn{2}{c}{Error $(\%)$ (p-value-T/p-value-W)} \\
\cline{2-3}
 & Sigmoid & ReLU \\
\hline
\hline
No dropout & 1.58 $\pm$ 0.045 (6.5e-26/9.1e-7) & 1.15 $\pm$ 0.036 (1.7e-19/9.1e-7) \\
\hline
Bernoulli dropout & 1.35 $\pm$ 0.049 (3.1e-17/9.1e-7) & 1.06 $\pm$ 0.037 (3.6e-17/9.1e-7) \\
\hline
Adaptive dropout & 1.30 $\pm$ 0.072 (7.4e-11/1.1e-6) & 1.02 $\pm$ 0.027 (1.9e-11/1.2e-6) \\
\hline
DropConnect & 1.37 $\pm$ 0.058 (8.7e-19/9.1e-7) & 1.01 $\pm$ 0.052 (8.5e-6/6.0e-5) \\
\hline
\hline
Uniform dropout & 1.21 $\pm$ 0.046 (9.0e-7/1.2e-5) & 0.96 $\pm$ 0.039 (0.031/0.027) \\
\hline
Gaussian dropout & \textbf{1.15 } $\pm$ 0.035 & \textbf{0.95} $\pm$ 0.028 \\
\hline
\end{tabular}
\end{center}
\end{table*}

\begin{figure*}
\begin{center}
%\vspace{-6mm}
    \centering
    \subfigure[\label{test.vs.epoch.confmat:1}]{\includegraphics[width=0.45\textwidth]{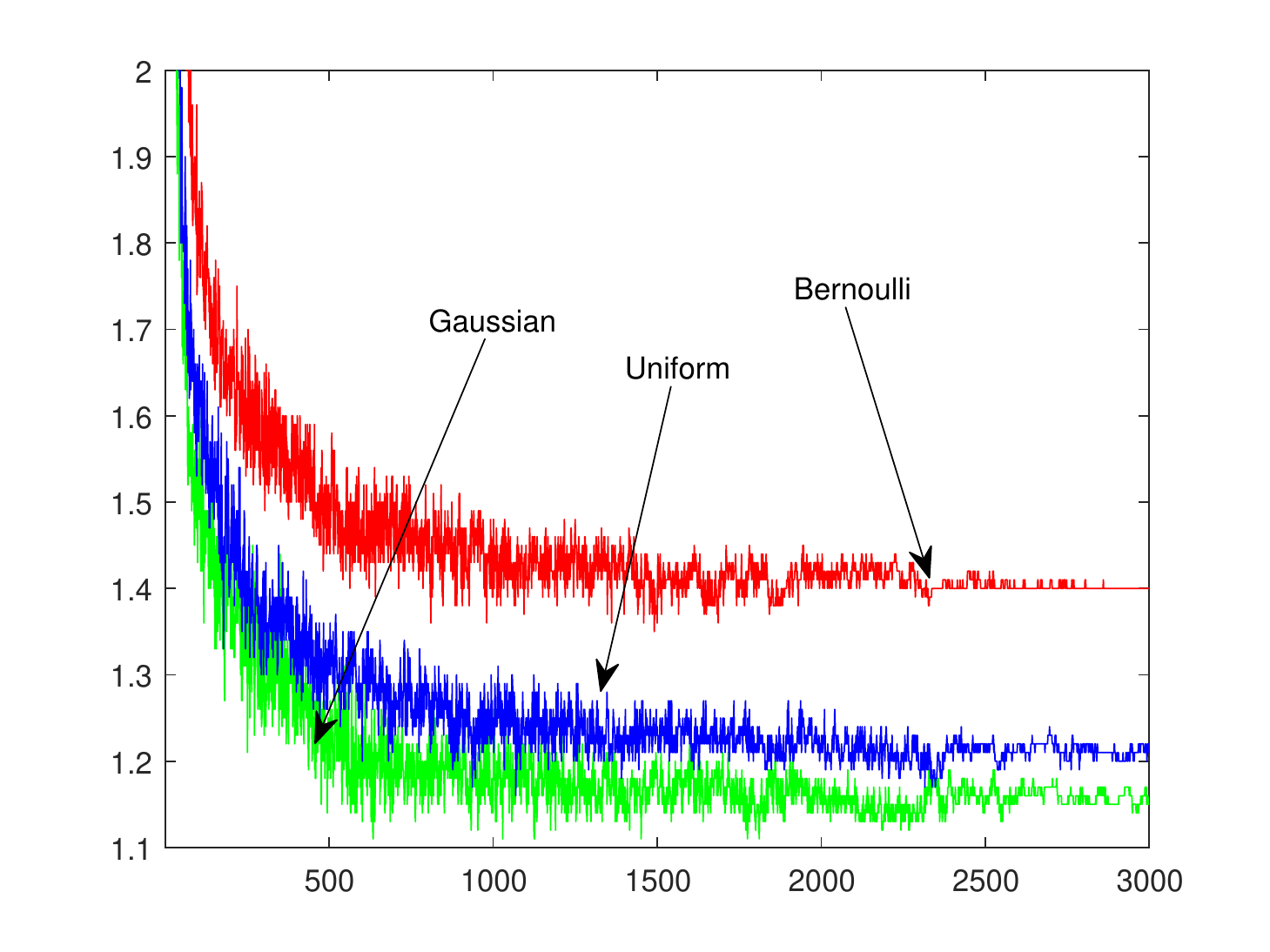}}
    \subfigure[\label{test.vs.epoch.confmat:2}]{\includegraphics[width=0.45\textwidth]{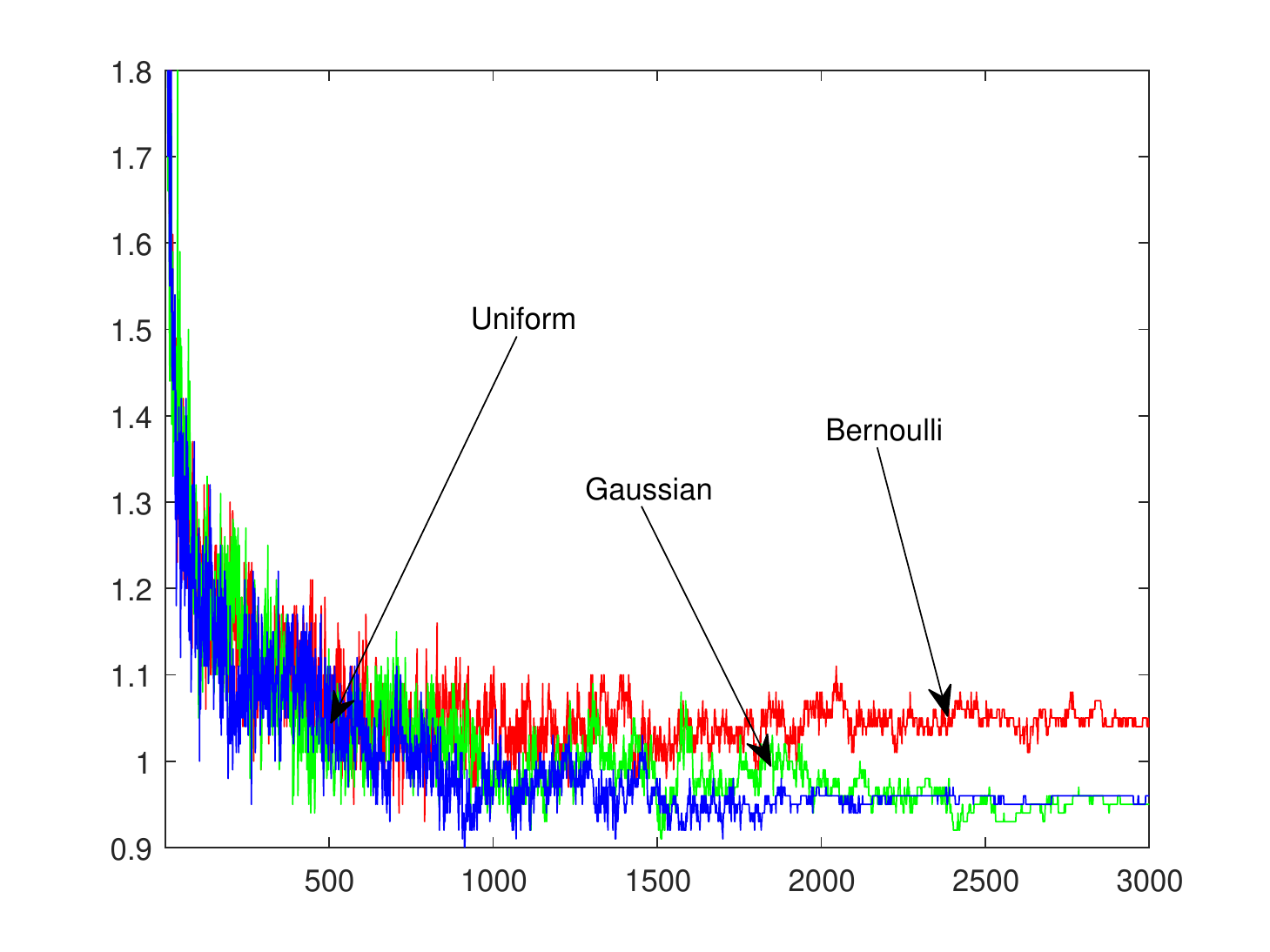}}
    \caption{ (a) and (b) show the \emph{testing errors vs. epochs} of Bernoulli, uniform,
        and Gaussian dropouts on MNIST. No data augmentation is used in this
        experiment. By regularizing the covariance between neurons
        in the same layer, the capacity of the neural network is improved. }
    \label{fig:test.vs.epoch}
\end{center}
\end{figure*}

\subsection{Experiments on MNIST}
We first verify the effectiveness of our continuous dropout on MNIST. The MNIST
handwritten digit dataset consists of $60000$ training images and $10000$ test
images. Each image is $28 \times 28$ pixels in size. We randomly separate the
$60000$ training images into two parts: $50000$ for training and $10000$ for
validation. We replicate the results of dropout in \cite{hinton1} and
use the same settings for uniform dropout and Gaussian dropout. These settings include a
linear momentum schedule, a constant weight constraint, and an exponentially
decaying learning rate. More details can be found in \cite{hinton1}.

We train models with two fully connected layers using sigmoid or ReLU
activation functions $(784-800-800-10)$.
%Specifically, in Gaussian Dropout, we set $\sigma=0.2$.
Table I shows the performance when
image pixels are taken as the input and no data augmentation is utilized. From this
table, we can see that both uniform dropout and Gaussian dropout outperform
Bernoulli dropout, adaptive dropout, and DropConnect on this dataset, irrespective of whether
Sigmoid or ReLU is applied. Gaussian dropout achieves slightly better
performance than uniform dropout. To further analyze the effects of continuous
dropout, Fig. \ref{fig:test.vs.epoch} shows the testing errors vs. epochs of Bernoulli dropout,
uniform dropout, and Gaussian dropout. We can see that continuous dropout
achieves a considerably lower testing error than Bernoulli dropout, which demonstrates that
continuous dropout has a better generalization capability.

%\begin{figure}[htb!f]
\begin{figure}[htb!]
\begin{center}
\includegraphics[width=0.47\textwidth]{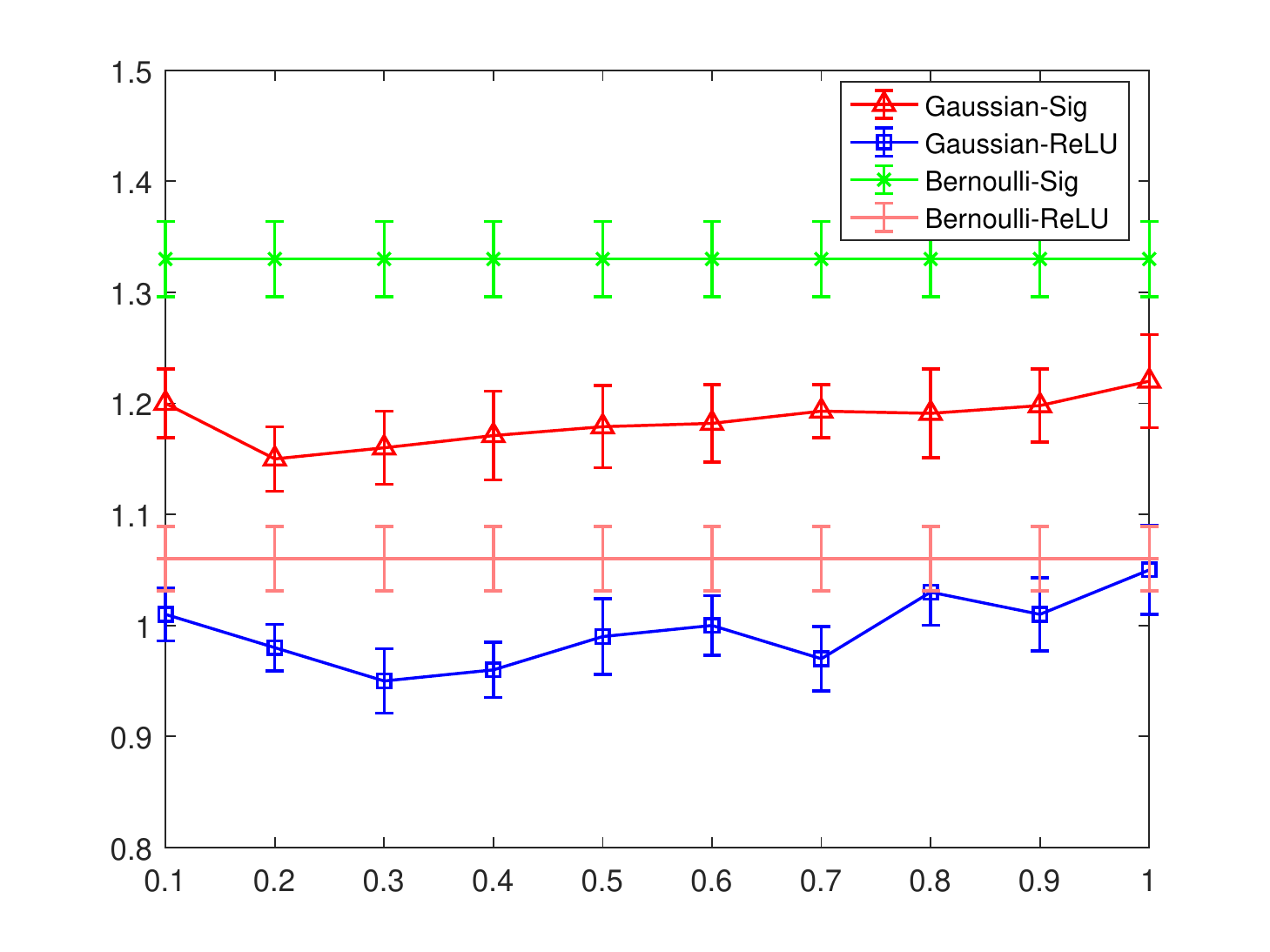}
\caption{ Performance curve of Gaussian Dropout w.r.t different variance. Our Gaussian Dropout consistently outperforms
Bernoulli Dropout for all sigma values. It shows that the performance gain in Gaussian Dropout mainly comes from the distribution not the extra freedom of $\sigma$.}
\label{fig:variance}
\end{center}
\end{figure}

\begin{figure*}[!t]
	\centering
    \subfigure[\label{convhist.confmat:1}]{\includegraphics[width=0.45\textwidth]{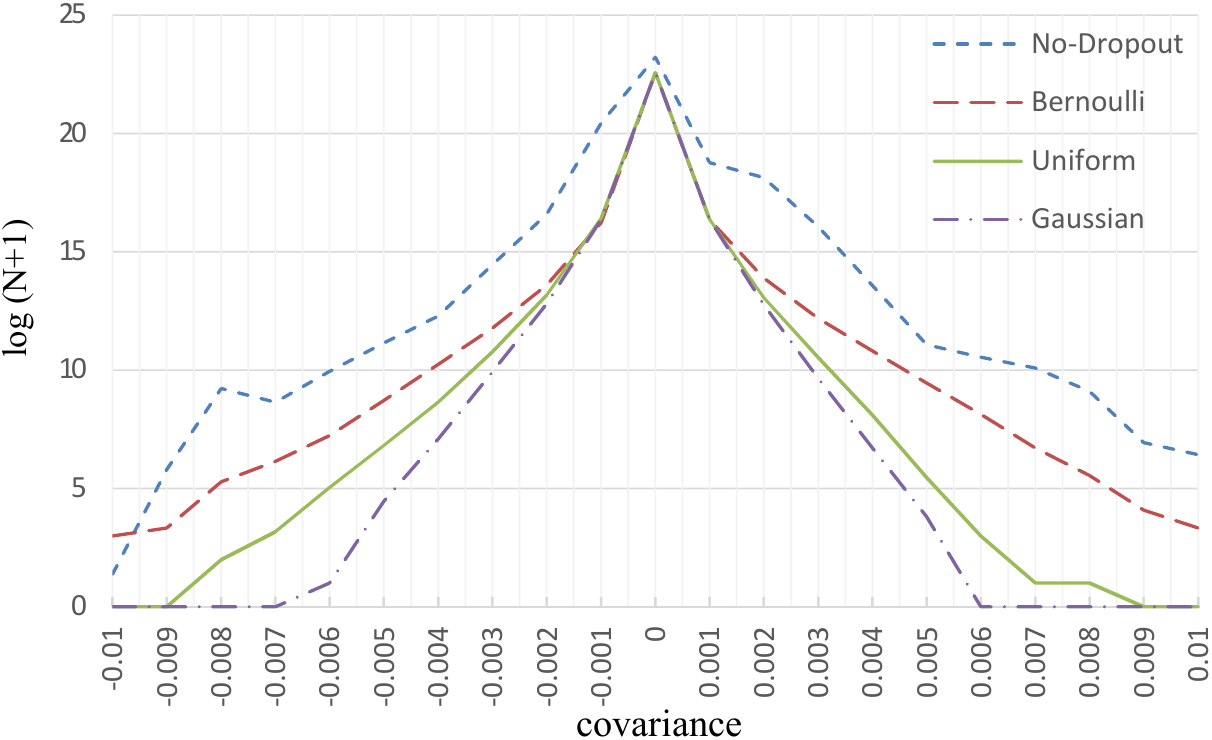}}
    \subfigure[\label{convhist.confmat:2}]{\includegraphics[width=0.45\textwidth]{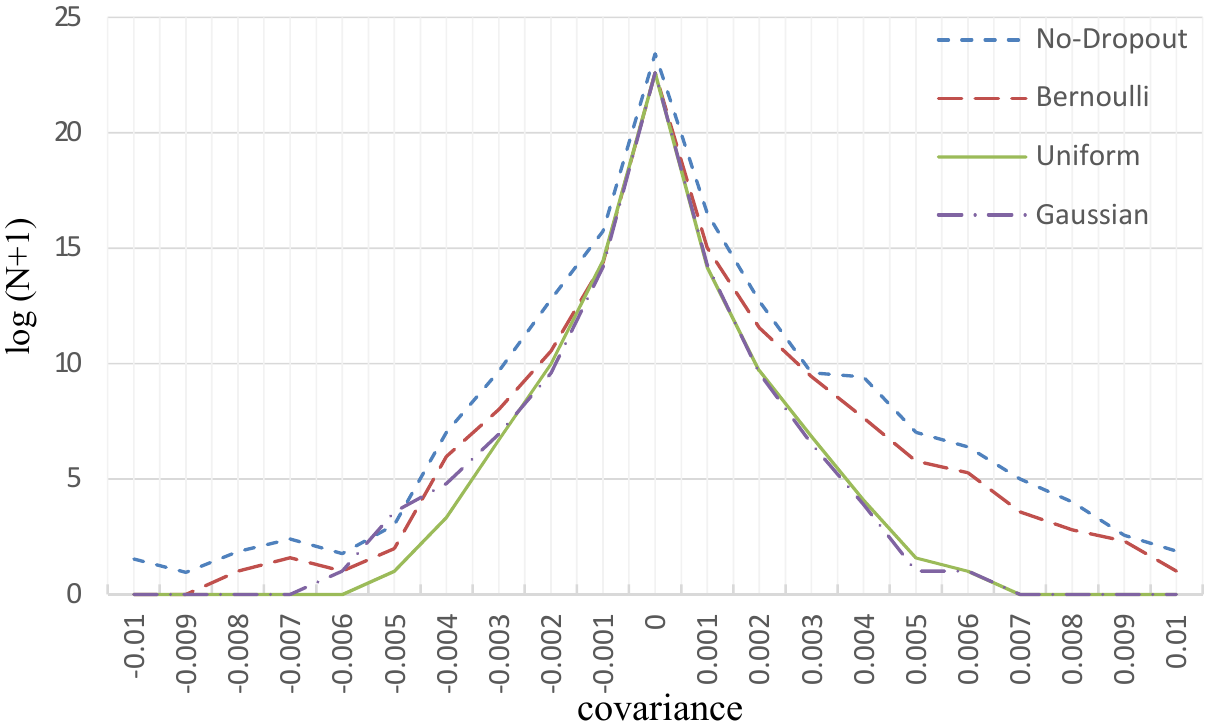}}
    \caption{ Log histogram of covariance between pairs of units from the same layer. Left: layer $1$; right: layer $2$. It shows that in continuous dropout, the
distribution is more concentrated around $0$, which indicates that continuous dropout performs better than Bernoulli dropout in preventing the co-adaptation
of feature detectors. (MNIST, $784-800-800-10$, ReLU)}
    \label{fig:covhist}
\end{figure*}

%\begin{figure*}[htb!f]
%\begin{center}
%\includegraphics[width=1\textwidth]{figs/34.png}
%%\vspace{-6mm}
%\caption{ Log histogram of covariance between pairs of units from the same layer. Left: layer $1$; right: layer $2$. (MNIST, $784-800-800-10$, ReLU)}
%\label{fig:covariance}
%\end{center}
%\end{figure*}

\begin{table*}[t]
\label{tab:minist-gau}
\caption{Performance comparison on MNIST with Gaussian Initialization (mean error and standard derivation). No data augmentation is used. Paired t-test and paired Wilcoxon signed rank test are conducted between Gaussian dropout and all other baseline methods. Their p-values are reported: p-value-T for t-test and p-value-W for Wilcoxon signed rank test.}
\begin{center}
\begin{tabular}{l|c|c|c}
\hline
Method & Architecture & Act Function & Error(\%) (p-value-T/p-value-W) \\
\hline
\hline
No dropout & 2CNN+1FC & ReLU & 0.674 $\pm$ 0.047 (1.8e-15/9.1e-7) \\
\hline
Bernoulli dropout & 2CNN+1FC & ReLU & 0.551 $\pm$ 0.017 (3.3e-5/1.5e-4) \\
\hline
Adaptive dropout & 2CNN+1FC & ReLU & 0.591 $\pm$ 0.017 (8.6e-18/9.1e-7) \\
\hline
DropConnect & 2CNN+1FC & ReLU & 0.581 $\pm$ 0.012 (4.2e-18/9.1e-7) \\
\hline
\hline
Uniform dropout & 2CNN+1FC & ReLU & 0.549 $\pm$ 0.021 (2.9e-3/4.5e-3) \\
\hline
Gaussian dropout & 2CNN+1FC & ReLU & \textbf{0.534} $\pm$ 0.006 \\
\hline
\end{tabular}
\end{center}
\end{table*}

\begin{table*}[t]
\label{tab:minist-uni}
\caption{Performance comparison on MNIST with Uniform Initialization (mean error and standard derivation). No data augmentation is used. Paired t-test and paired Wilcoxon signed rank test are conducted between Gaussian dropout and all other baseline methods. Their p-values are reported: p-value-T for t-test and p-value-W for Wilcoxon signed rank test.}
\begin{center}
\begin{tabular}{l|c|c|c}
\hline
Method & Architecture & Act Function & Error(\%) (p-value-T/p-value-W) \\
\hline
\hline
No dropout & 2CNN+1FC & ReLU & 0.670 $\pm$ 0.051 (3.0e-16/9.1e-7) \\
\hline
Bernoulli dropout & 2CNN+1FC & ReLU & 0.558 $\pm$ 0.018 (4.4e-8/6.5e-6) \\
\hline
Adaptive dropout & 2CNN+1FC & ReLU & 0.586 $\pm$ 0.016 (3.2e-13/1.0e-6) \\
\hline
DropConnect & 2CNN+1FC & ReLU & 0.579 $\pm$ 0.011 (4.6e-16/9.1e-7) \\
\hline
\hline
Uniform dropout & 2CNN+1FC & ReLU & 0.558 $\pm$ 0.018 (1.3e-11/1.4e-6) \\
\hline
Gaussian dropout & 2CNN+1FC & ReLU & \textbf{0.521} $\pm$ 0.017 \\
\hline
\end{tabular}
\end{center}
\end{table*}

\textbf{Influence of variance in Gaussian Dropout}. 
In Section III, we find
that we have an extra degree of freedom by using $\sigma^2$ to achieve the balance between
network output and model complexity. To investigate the influence of $\sigma^2$ on model
performance in Gaussian Dropout, we train Gaussian Dropout models with $784-800-800-10$ neurons.
Dropout masks are sampled from Gaussian distribution with mean $0.5$ and variance
in $\{0.1, 0.2, 0.3, 0.4, 0.5, 0.6, 0.7, 0.8, 0.9, 1.0\}$. Activation functions are set to be
Sigmoid or ReLU. Performance of Gaussian Dropout with different standard deviations are shown in
Fig. \ref{fig:variance}. We can see that the best variance for Gaussian Dropout is $\{0.2, 0.3\}$.
For normal distributions, the values less than two standard deviations from the mean account for
$95.45\%$ of the set. And for three standard deviations, that is $99.73\%$.
Thus, almost all the values of $\mathcal{N}(0.5, 0.2)$ and $\mathcal{N}(0.5, 0.3)$ distribute in
$[0,1]$ (reasonable distribution for dropout mask variables).
Most importantly, our Gaussian Dropout consistently outperforms
Bernoulli Dropout for all sigma values, which demonstrate that the performance gain in Gaussian Dropout mainly comes from the distribution not the
extra freedom of sigma.

\textbf{Covariance of hidden units}. In Section III, we demonstrate that
continuous dropout can prevent the co-adaptation of feature detectors. To verify
this property, we investigate the distribution of covariance between units in
the same layer. We construct histograms of the variance of all pairs of units in
the same layer in a trained $784-800-800-10$ MNIST model with ReLU. Figure \ref{fig:covhist}
shows the log of the number of pairs $(N)$ whose covariance falls into different
intervals. Histograms are obtained by taking all the $800 \times 800$ unit
pairs in each layer and aggregating the results over $10$ random input samples.
For each sample, the dropout process is repeated $10000$ times to estimate the
covariance. Figure \ref{fig:covhist} shows that in continuous dropout, the
distribution is more concentrated around $0$, which indicates that continuous
dropout performs better than Bernoulli dropout in preventing the co-adaptation
of feature detectors.
Furthermore, comparing Fig. \ref{fig:covhist}(a) and Fig. \ref{fig:covhist}(b), we can see that in ``No dropout'' the covariance in second layer is much more concentrated around 0 than that in the first layer. After using Continuous Dropout, the covariance curve becomes more concentrated than ``No dropout'' in both layers. The reason why the effects of Continuous Dropout become less significant in a higher layer is that the room for improvement (reduce covariance) becomes smaller in a higher layer.

%\textcolor{red}{Furthermore, comparing
%    \ref{fig:covhist}(a) with (b), we can see that even for non-dropout
%network, the convariances of layer $2$ are generally less than layer $1$, so
%the reduce of covariances for deeper layers comes from the deep architectures
%and not from the usage of dropout}.

To further improve the classification results, we also apply a more powerful
network, which consists of a $2$-layer CNN with $32-64$ feature maps and $1$ fully
connected layer with $150$ ReLU units. All the dropout algorithms are applied
on the fully connected layer. We use an initial learning rate of $0.01$ and
manually decay the learning rate by a multiplier ($0.5$ or $0.1$) when
the loss function of the validation error reaches a plateau. The input is also the
original image pixels without cropping, rotation, or scaling.
To verify whether the
improvement of continuous dropout is benefited from a favoured initialization,
we initialize weights by using both Gaussian distribution ($\mathcal{N}(0, 0.01)$)
and Uniform distribution proposed by Glorot and Bengio \cite{xavier-filler}.
The experimental results are summarized in Table II and Table III. We can see that Gaussian dropout
consistently performs the best among all dropout methods, no matter which initialization distribution is applied.
Paired t-test and paired Wilcoxon signed rank test are conducted between Gaussian dropout and other methods. Both tables show that Gaussian
dropout achieves statistically significant improvement over all baseline methods and the p-values are less than 0.05.

%From the t-test point of view, the improvement achieved by Gaussian dropout is not as significant as in the
%$782-800-800-10$ model. This is because the $782-800-800-10$ model is with a
%small complexity and faced with a underfitting case in training (Fig. 2), Gausian
%dropout improves the capacity of the model by pushing the network to learn more independent neurons (features).
%While in this CNN network, the model is complex enough to fitting the mapping
%between the input and output, some degree of redundancy in neurons do not
%result in a significant performance degradation.

\subsection{Experiments on CIFAR-10}
The CIFAR-$10$ dataset consists of $10$ classes of $32 \times 32$ RGB images with
$50000$ for training and $10000$ for testing. We preprocess the data by global
contrast normalization and ZCA whitening as in \cite{Good23}. To
produce comparable results to the state-of-the-art method, we apply all the dropout
algorithms on the Network In Network (NIN) model \cite{Lin22}. This network
consists of $7$ convolutional layers and part of them are connected to pooling
layers. Two dropout layers are applied to the pooling layers. To
compare continuous dropout with adaptive dropout and DropConnect, we slightly
change this model by omitting the two dropout layers between the CNNs and
replace the last pooling layer by two fully connected layers with $128$ and
$10$ units, respectively. Dropout is applied to the first fully connected
layer. During training, we first initialize our model by the weights trained in
\cite{Lin22}, and then we finetune the model using different dropout
methods. The learning rate is initialized by $0.01$ and decayed by $10$ every $3000$ iterations,
without any data augmentations.

\begin{table}[t]
\label{perf:CIFAR}
\caption{Performance comparison on CIFAR-10 (mean error and standard derivation). Paired t-test and paired Wilcoxon signed rank test are conducted between Gaussian dropout and all other baseline methods. Their p-values are reported: p-value-T for t-test and p-value-W for Wilcoxon signed rank test.}
\begin{center}
\begin{tabular}{l|c}
\hline
Method   & Error(\%) (p-value-T/p-value-W) \\
\hline
\hline
No dropout & 10.65 $\pm$ 0.114 (1.5e-14/9.1e-7) \\
\hline
Bernoulli dropout  & 10.55 $\pm$ 0.050 (5.5e-15/9.1e-7) \\
\hline
Adaptive dropout & 10.46 $\pm$ 0.081 (1.6e-10/9.1e-7) \\
\hline
DropConnect  & 10.40 $\pm$ 0.178 (2.9e-7/7.8e-6)\\
\hline
\hline
Uniform dropout  & 10.47 $\pm$ 0.142 (5.2e-10/2.0e-6) \\
\hline
Gaussian dropout & \textbf{10.18} $\pm$ 0.129 \\
\hline
\end{tabular}
\end{center}
\end{table}

\begin{figure*}[!t]
	\centering
    \subfigure[\label{confmat.confmat:1}]{\includegraphics[width=0.40\textwidth]{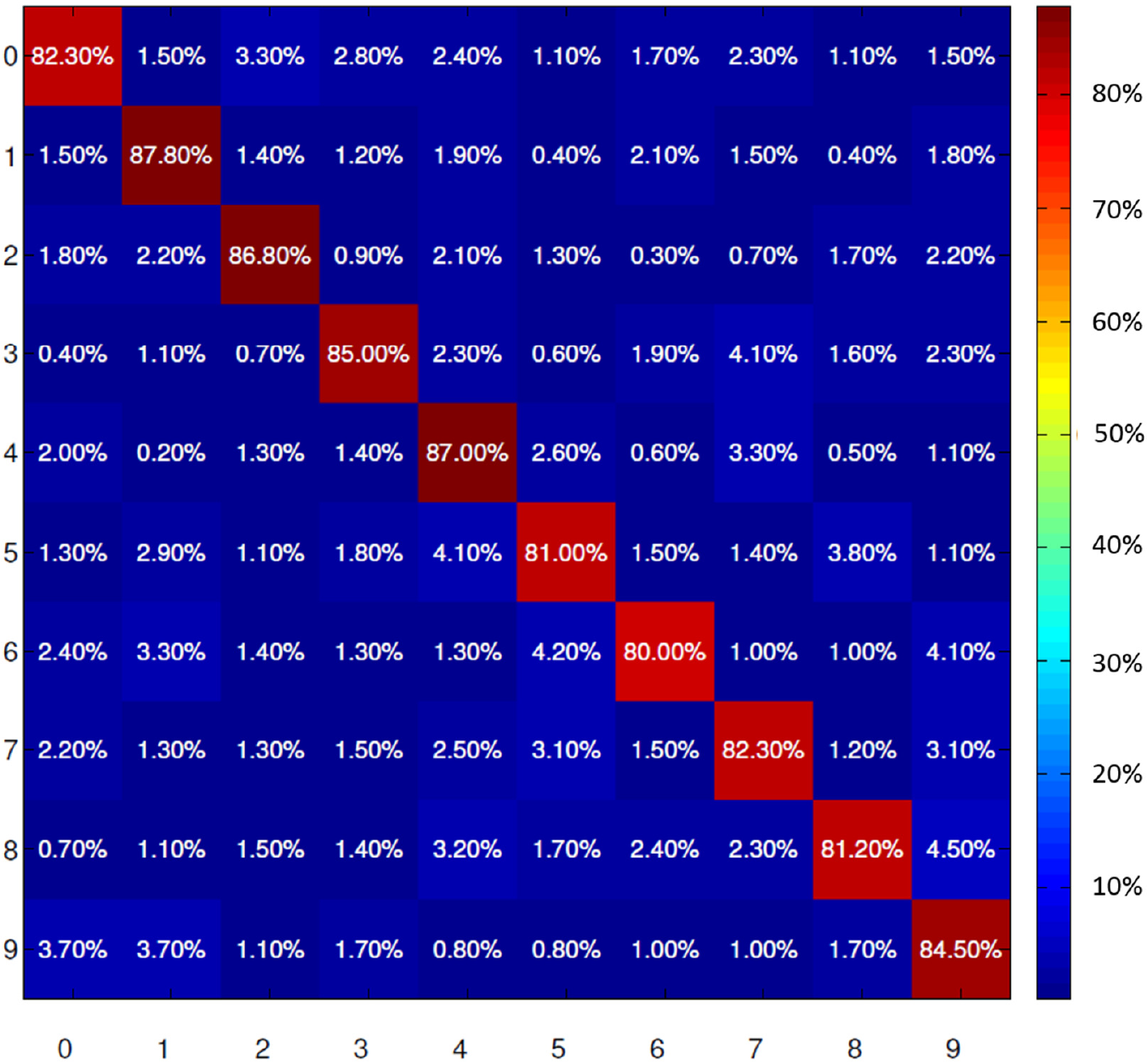}} \hspace{5mm}
    \subfigure[\label{confmat.confmat:2}]{\includegraphics[width=0.40\textwidth]{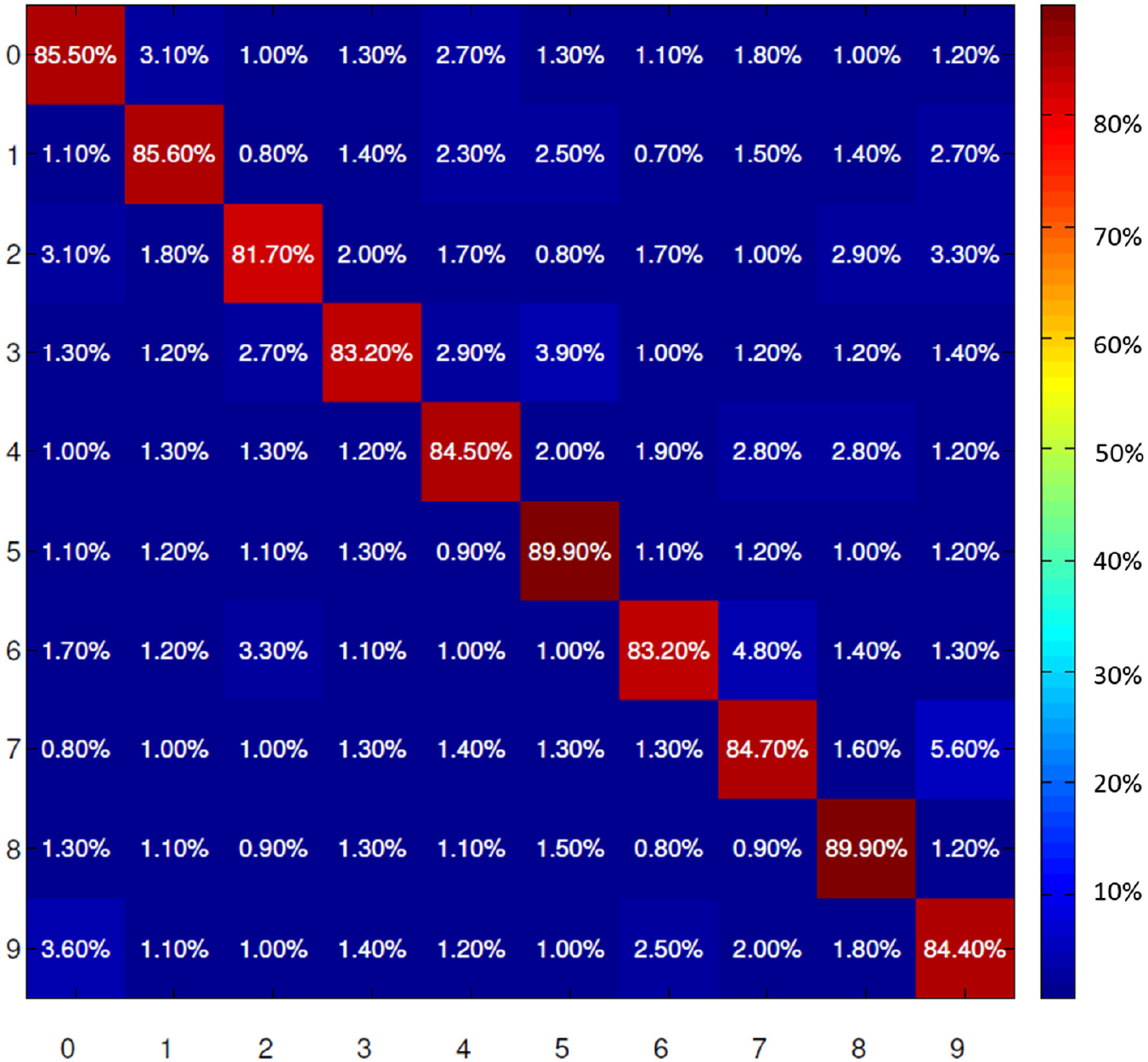}}\\
	\subfigure[\label{confmat.confmat:3}]{\includegraphics[width=0.40\textwidth]{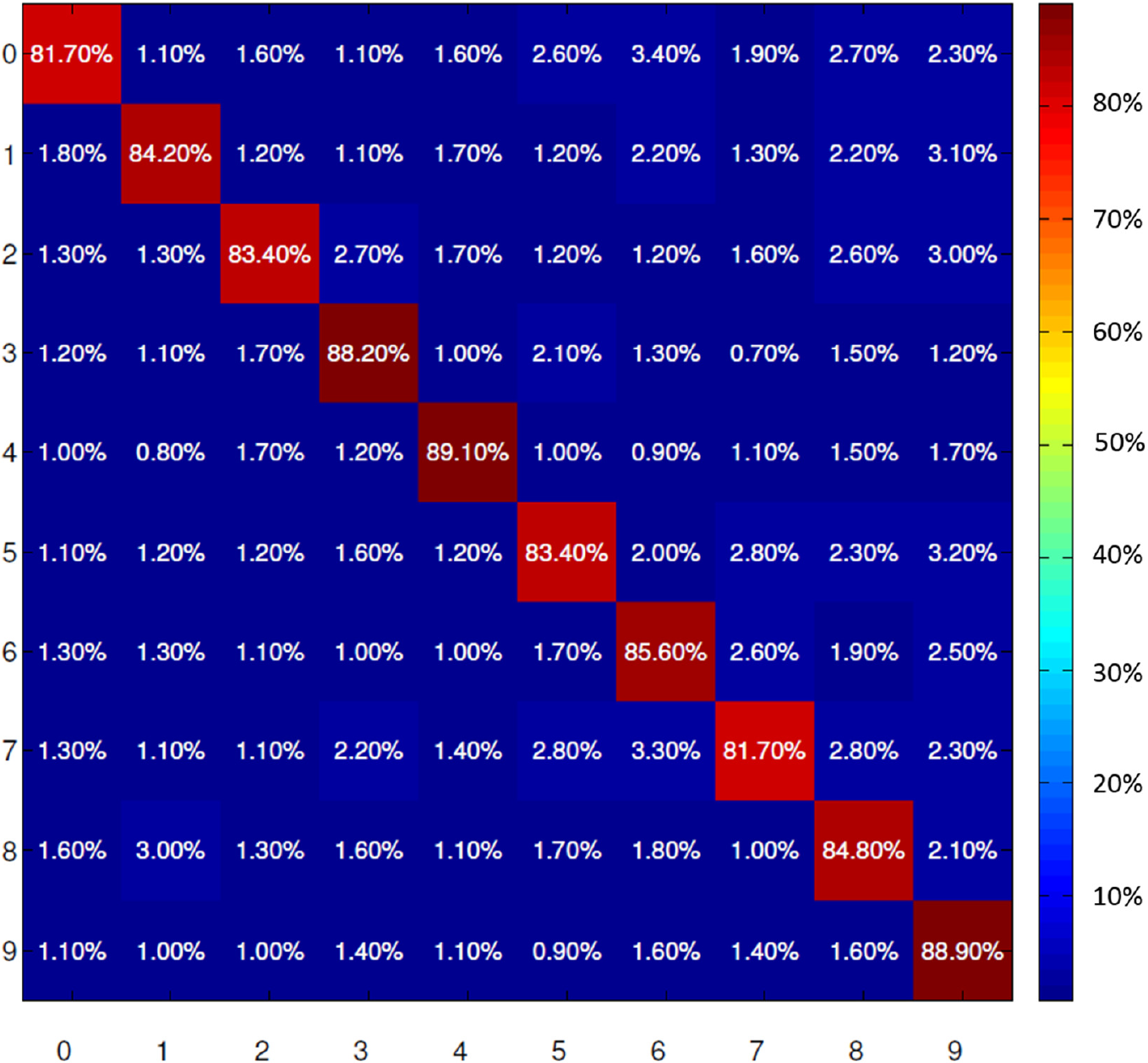}} \hspace{5mm}
    \subfigure[\label{confmat.confmat:4}]{\includegraphics[width=0.40\textwidth]{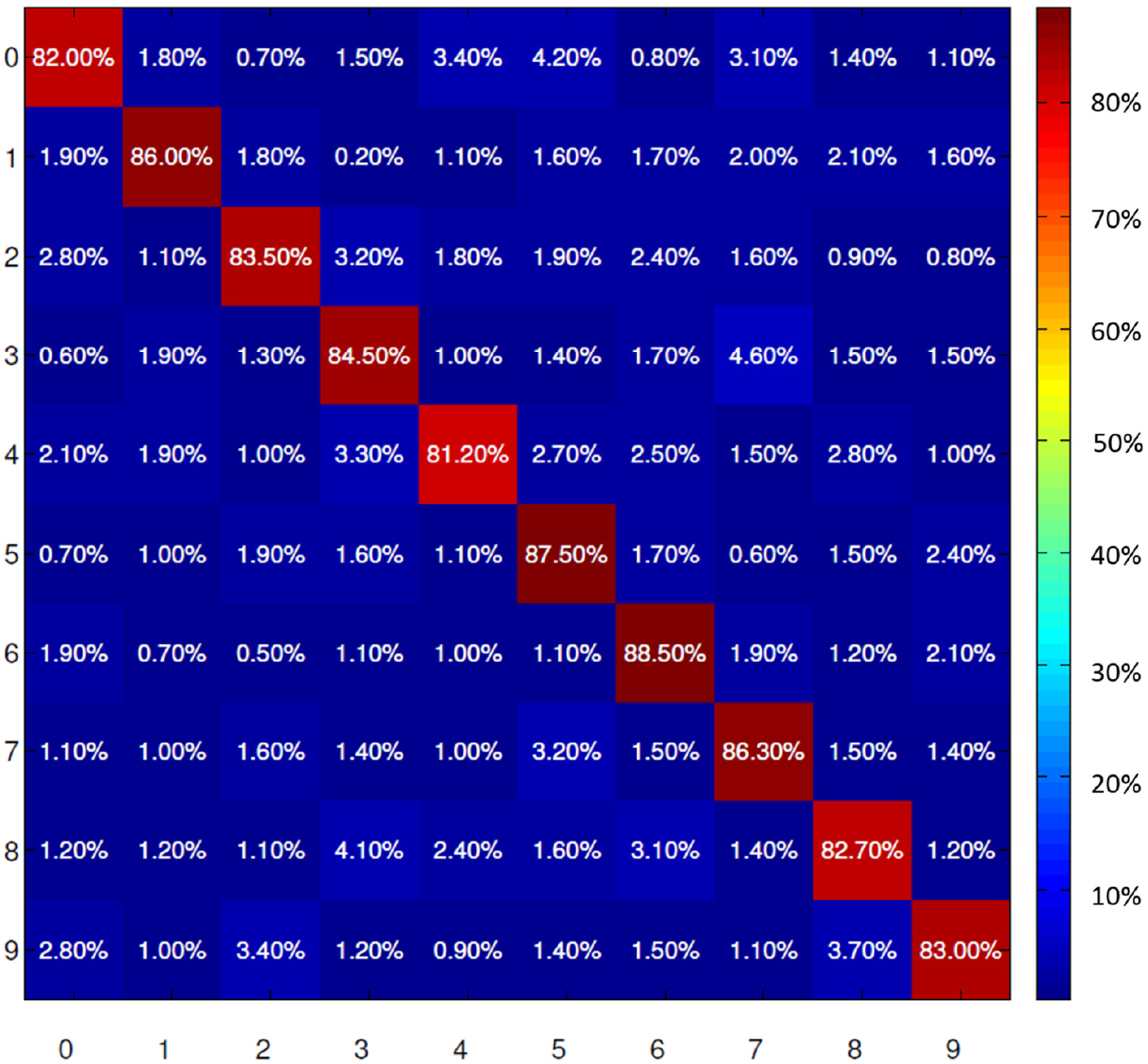}}\\
    \subfigure[\label{confmat.confmat:5}]{\includegraphics[width=0.40\textwidth]{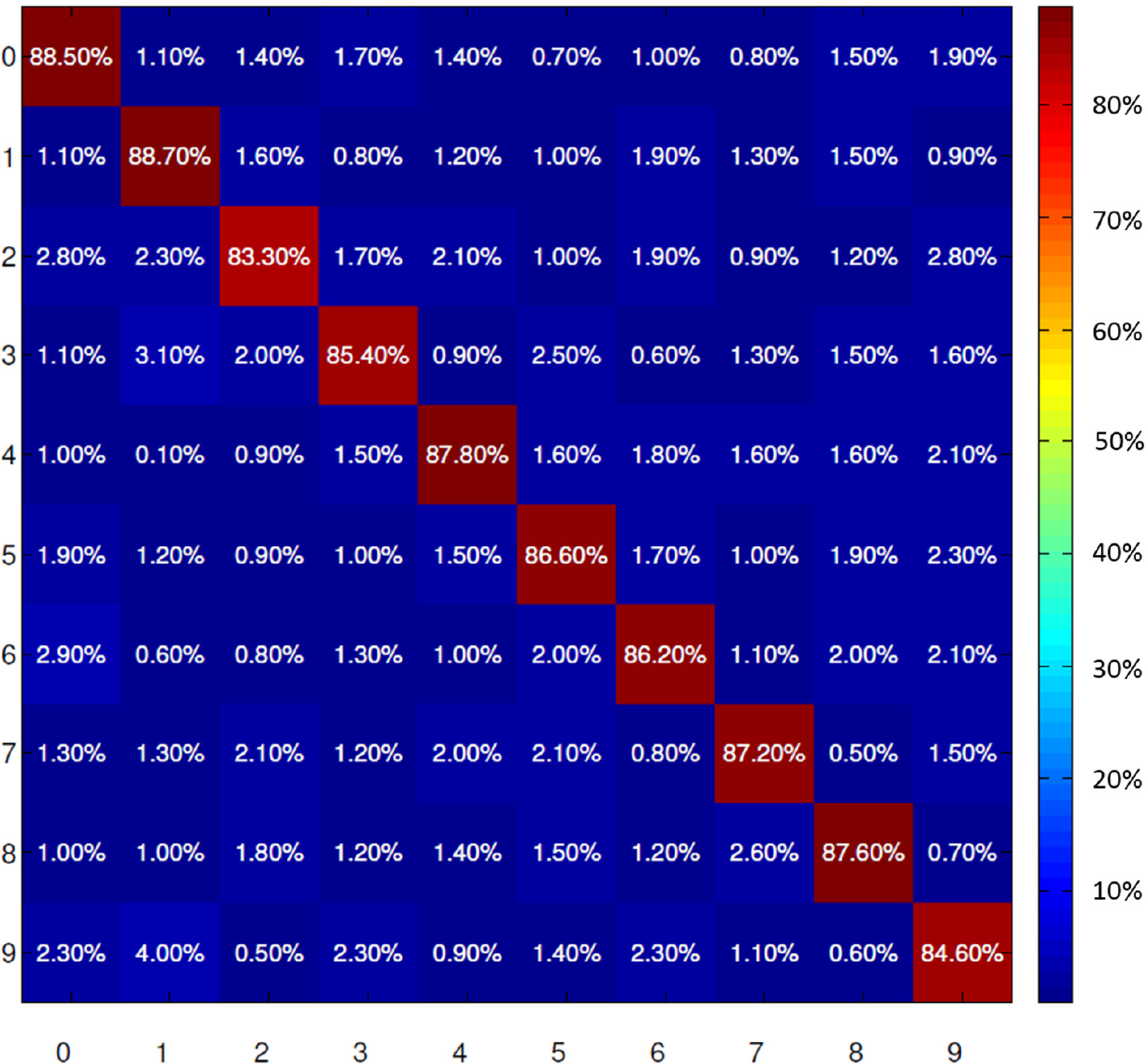}} \hspace{5mm}
    \subfigure[\label{confmat.confmat:6}]{\includegraphics[width=0.40\textwidth]{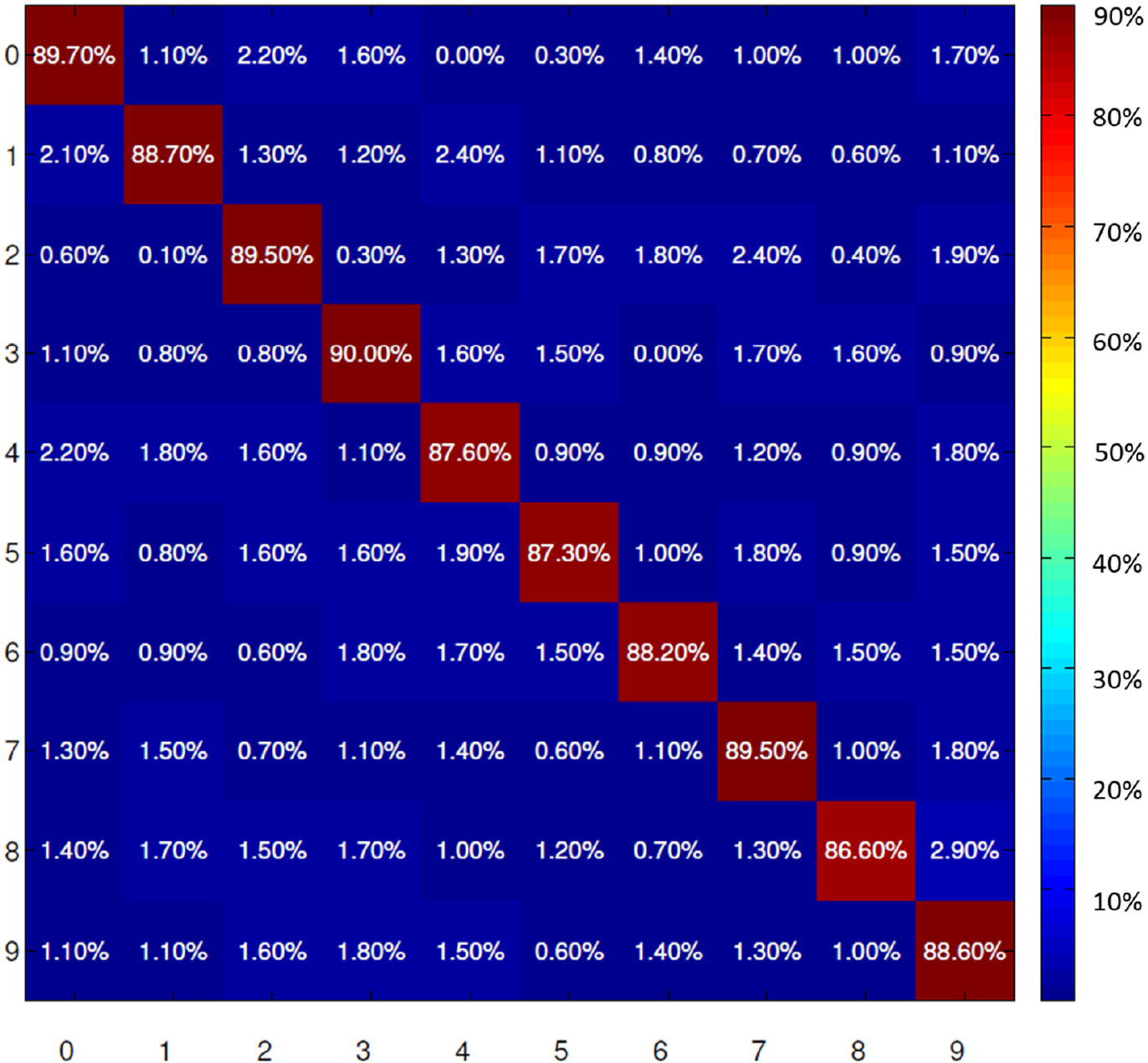}}\\
    \caption{The confusion matrices of all six methods: (a) no Dropout; (b) Bernoulli Dropout; (c) Adaptive Dropout; (d) DropConnect; (e) Uniform Dropout; (f)
    Gaussian Dropout. We can see that Gaussian Dropout achieves the best performance on five classes among all six methods. Specifically, Gaussian Dropout achieves higher classification accuracy on 10, 8, 8, 8, and 7 classes than No dropout, Bernoulli dropout, Adaptive dropout, DropConnect and Uniform dropout, respectively.}
    \label{fig:confmat}
\end{figure*}

The models are tested after $10000$ iterations, and the results are presented in Table
IV. We can see that Gaussian dropout achieves the best performance among all
dropout algorithms on this task again. Based on the results of paired t-test and paired Wilcoxon signed rank test,
Gaussian dropout significantly outperforms all other methods (p-values are less than 0.05).
To further investigate their performance on each class,
confusion matrices are also reported, as shown in Fig. \ref{fig:confmat}. We can see that Gaussian Dropout achieves the
best performance on five classes among all six methods. Specifically, Gaussian Dropout achieves higher classification accuracy
on 10, 8, 8, 8, and 7 classes than No dropout, Bernoulli dropout, Adaptive dropout, DropConnect and Uniform dropout, respectively.

%of this $10$ classes task in Fig. \ref{fig:confmat}, which
%indicates that our dropout has more consistant behavior for each class (diagonal elements)
%and more centralized mis-classfications (non-diagnal elements).

\begin{table}[t]
\label{perf:SVHN}
\caption{Performance comparison on SVHN (mean error and standard derivation). Paired t-test and paired Wilcoxon signed rank test are conducted between Gaussian dropout and all other baseline methods. Their p-values are reported: p-value-T for t-test and p-value-W for Wilcoxon signed rank test.}
\begin{center}
\begin{tabular}{l|c}
\hline
Method   & Error(\%) (p-value-T/p-value-W) \\
\hline
\hline
No dropout & 2.09 $\pm$ 0.002 (1.8e-35/9.1e-7) \\
\hline
Bernoulli dropout  & 2.00 $\pm$ 0.012 (4.2e-19/9.1e-7) \\
\hline
Adaptive dropout & 2.13 $\pm$ 0.235 (3.5e-5/1.3e-4) \\
\hline
DropConnect  & 1.96 $\pm$ 0.007 (8.4e-16/9.1e-7) \\
\hline
\hline
Uniform dropout  & \textbf{1.92} $\pm$ 0.012 (0.982/0.981) \\
\hline
Gaussian dropout & 1.93 $\pm$ 0.012 \\
\hline
\end{tabular}
\end{center}
\end{table}

\subsection{Experiments on SVHN}

The Street View House Numbers (SVHN) dataset includes 604388 training images (both
training set and extra set) and 26032 testing images \cite{SVHN}. Like MNIST, the goal
is to classify the digit centered in each $32\times32$ image ($0\sim9$). The dataset is augmented
by: 1) randomly select a $28\times28$ region from the original image; 2) introducing $15\%$ scaling
and rotation variations; and 3) randomly flip images during training. Following \cite{wan14},
we preprocess the images using local contrast normalization as in \cite{LCN}.

\begin{table}[t]
\label{perf:NORB}
\caption{Performance comparison on NORB (mean error and standard derivation). Paired t-test and paired Wilcoxon signed rank test are conducted between Gaussian dropout and all other baseline methods. Their p-values are reported: p-value-T for t-test and p-value-W for Wilcoxon signed rank test.}
\begin{center}
\begin{tabular}{l|c}
\hline
Method   & Error(\%) (p-value-T/p-value-W) \\
\hline
\hline
No dropout & 3.55 $\pm$ 0.070 (1.4e-16/9.1e-7) \\
\hline
Bernoulli dropout  & 3.33 $\pm$ 0.128 (1.6e-5/9.9e-5) \\
\hline
Adaptive dropout & 3.49 $\pm$ 0.204 (1.4e-8/2.0e-6) \\
\hline
DropConnect  & 3.53 $\pm$ 0.059 (1.2e-15/9.1e-7) \\
\hline
\hline
Uniform dropout  & 3.29 $\pm$ 0.197 (1.2e-3/2.4e-3) \\
\hline
Gaussian dropout & \textbf{3.15} $\pm$ 0.114 \\
\hline
\end{tabular}
\end{center}
\end{table}

\makeatletter\def\@captype{table}\makeatother
\begin{table*}[htbp]
\centering
\caption{Performance Comparison on ImageNet ILSVRC-2012 (mean top-5/top-1 error and standard derivation). Paired t-test and paired Wilcoxon signed rank test are conducted between Gaussian dropout and all other baseline methods. Their p-values are reported: p-value-T for t-test and p-value-W for Wilcoxon signed rank test.}
\label{perf:ImageNet}
\begin{tabular*}{1.0\textwidth}{c|c|c|c|c|c}
\cline{1-6}
\multirow{2}{*}{Method} & \multirow{2}{*}{ConvNet config.} & \multicolumn{2}{c| }{smallest image side} & \multirow{2}{*}{top-5 error(\%) (p-value-T/p-value-W)} & \multirow{2}{*}{top-1 error(\%) (p-value-T/p-value-W)} \\
\cline{3-4}& & train(S) & test(Q) & & \\
\cline{1-6}
 Bernoulli Dropout\cite{VGG} & \multirow{5}{*}{VGG\_ILSVRC\_16\_layers} &
 \multirow{5}{*}{256} & \multirow{5}{*}{256} & 8.86 $\pm$ 0.042 (9.5e-11/9.8e-4) & 26.99 $\pm$ 0.065 (7.4e-11/9.8e-4) \\
\cline{1-1}\cline{5-6} Adaptive Dropout &  &  &  & 8.41 $\pm$ 0.061 (1.1e-6/9.8e-4)& 26.27 $\pm$ 0.046 (2.8e-8/9.8e-4) \\
\cline{1-1}\cline{5-6} DropConnect &  &  &  & 8.56 $\pm$ 0.037 (2.3e-8/9.8e-4)& 26.82 $\pm$ 0.050 (6.2e-11/9.8e-4) \\
\cline{1-1}\cline{5-6} Uniform Dropout &  &  &  & 8.08 $\pm$ 0.048 (0.017/0.024) & 25.91 $\pm$ 0.046 (0.005/0.014)\\
\cline{1-1}\cline{5-6} Gaussian Dropout &  &  &  & \textbf{7.99} $\pm$ 0.065 & \textbf{25.79} $\pm$ 0.045 \\
\cline{1-6}
\end{tabular*}
\end{table*}

%\begin{figure}
%\begin{center}
%    \centering
%    \includegraphics[width=0.15\textwidth]{figs/svhn_norb_network}
%    \caption{Baseline architecture for SVHN and NORB. Convolution layers are
%    shown in red, pooling layers in blue, fully connected layers in orange. The
%    filter sizes are indicated in the right, number of filters in the left.
%    ReLU are used for activation. Dropout is applied on the first two fully connected
%    layers.}
%    \label{fig:svhn_norb_network}
%\end{center}
%\end{figure}

%Following \cite{dieleman-cyclic-2016}, we use a VGG inspired architecture for
%this dataset by using $3\times3$ `same' convolutions throughout in combination
%with (overlapping) pooling. It is shown in Fig. \ref{fig:svhn_norb_network}.
The model consists of 2 convolutional layers and 2 locally connected layers as described in \cite{Kriz21}
(layers-conv-local-11pct.cfg). A fully connected layer with $512$ neurons and ReLU activations
is added between the softmax layer and the final locally connected layer.
We manually decrease the learning rate if the performance on validation set goes to plateaus \cite{Kriz21}. In detail, we multiply
the initial learning by $0.5$ and then $0.1$ repeatedly. Initial learning rate is set to $0.01$.
The bias learning rate is set to be $2\times$ the learning rate for the weights.
Additionally, weights are initialized with $\mathcal{N}(0, 0.1)$ random values for fully connected layers
and $\mathcal{N}(0, 0.01)$ for convolutional layers. To further improve the performance, we train $5$
independent networks with random permutations of the training sequence and different random seeds.
We report the classification error of averaging the output probabilities from the $5$ networks
before making a prediction.

The experimental results on this dataset are summarized in Table V. Comparing the mean classification errors, standard deviations,
and p-values of paired t-test and paired Wilcoxon signed rank test, we can see that our proposed continuous dropout achieves better performance than No dropout, Bernoulli dropout, Adaptive dropout, and DropConnect. The performance gain of Gaussian dropout is statistically significant (all p-values are less than 0.05). Uniform dropout and Gaussian dropout achieve similar performance on this dataset.
Besides, all dropout methods achieves stable performance on this dataset with small standard deviation,
except Adaptive dropout which has a large standard deviation (0.235).

%In this experiment, due to the large training set size, \textcolor{red}{our
%continuous dropout achieves the best performance with a significant gain.
%With a significant diversity in large number of training images, higher
%independence in feature extractors contributes to better generalization ability.}

\subsection{Experiments on NORB}
In this experiment we evaluate our models on the 2-fold NORB (jittered-cluttered) dataset \cite{NORB}.
Each image is classified into one of the six classes, which appears on a random background. Images are
downsampled from $108\times108$ to $48\times48$ as in \cite{downsample_norb}. We train on 2-fold of
29160 images each and test on a total of 58320 images.
We use the same architecture as in SVHN. Dataset is augmented by $15\%$ rotation and scaling.
No random crop or flip is applied. Models are trained with an initial learning
rate of $0.01$. Other training and testing settings are the same as in SVHN.

The experimental results are given in Table VI. From this table, we can see that Gaussian dropout significantly
outperforms No dropout, Adaptive dropout, DropConnect and Uniform Dropout on this dataset.
%However, when compared with Bernoulli dropout,
%the p-value of paired t-test is 0.081, which is larger than 0.05. It indicates that although Gaussian Dropout achieves
%smaller classification error than Bernoulli dropout, but the gain is not statistically significant.
Compared with the results on SVHN dataset, all methods have a larger standard deviation on NORB dataset.
Experiments on these two dataset adopt the same network architecture and other experimental settings.
The reason for higher standard deviation on NORB dataset may be that we have much fewer training images on NORB.
Therefore, the models trained on NORB are not as stable as that on SVHN.

%Note that our continuous dropout outperforms all other dropout algorithms again
%but still not significant. \textcolor{red}{The highly various
%    appearances in the large number of training samples ($583,200$) act as some
%kind of noise, which reduces the stability of trained model, so it's hard for a
%model to consistently outperform other models}.

\subsection{Experiments on ILSVRC-2012}
The ILSVRC-2012 dataset was used for ILSVRC $2012-2014$ challenges. This data set
includes images of $1000$ classes, and is split into three sets: training (1.3M images),
validation (50K images), and testing (100K images with held-out class labels). The classification
performance is evaluated using two measures: the top-1 and top-5 error. The former is a multi-class
classification error, and the latter is the main evaluation criteria used in ILSVRC, and is computed
as the proportion of images such that the ground-truth category is outside the top-5 predicted categories.

We compare all the dropout algorithms by finetuning on the model with $16$ layers proposed by VGG
team (configuration D) in \cite{VGG}. The model consists of $13$ convolution layers and $3$ fully connected
layers. All the filters used in the convolution layers are configured with $3\times3$ receptive field, and
number of channels are $\{64, 64, 128, 128, 256, 256, 256, 512, 512, 512, 512, 512, 512\}$ respectively.
The convolution stride is fixed to $1$ pixel; the spatial padding of convolution layer input is $1$ pixels
to preserve the spatial resolution of input. The convolutional layers are followed by three Fully-Connected (FC)
layers: the first two have $4096$ channels each, the third contains $1000$ channels to perform $1000$ way ILSVRC
classification. All hidden layers are equipped with the rectification (ReLU \cite{AlexNet}) and Bernoulli dropout is imposed
on the first two FC layers. In our experiment, the two FC layers with Bernoulli dropout are replaced by Adaptive dropout
FC layers, DropConnect FC layers, FC layers with Uniform dropout and Gaussian dropout, respectively.

During training, weights are first initialized by the VGG\_ILSVRC\_16\_layers
model\footnote{\url{http://www.robots.ox.ac.uk/~vgg/research/very_deep/}} in \cite{VGG},
then finetuned by $10,0000$ iterations. The input to the ConvNet is a fixed-sized $224\times224$ RGB images, which
are zero-centered by a subtraction of $[103.939, 116.779, 123.68]$ on BGR values.
The batch size was set to $64$, momentum to $0.9$, gradient clip to $35$. The finetuning was regularized by weight
decay $5\times 10^{-4}$. For Adaptive dropout, alpha is set to $1$,
beta is set to $0$. Dropout ratio is $0.5$ in DropConnect. In Uniform dropout, mask is sampling from $U[0,1]$. While the
Gaussian dropout mask is sampled from $\mathcal{N}(0.5, 0.3^2)$. The learning rate was initially set to $10^{-4}$,
and then decreased by a factor of $10$ after $50,000$ iterations. Following \cite{VGG},
the smallest side (denoted as S) of the training images are isotropically-rescaled to $256$.

\makeatletter\def\@captype{table}\makeatother
\begin{table*}[htbp]
\centering
\caption{Performance rank of different dropout methods on all five datasets.}
\label{perf:rank}
\begin{tabular*}{0.66\textwidth}{c|c|c|c|c|c|c}
\hline
Method & MNIST & CIFAR-10 & SVHN & NORB & ILSVRC-2012 & Average rank\\
\hline
Bernoulli Dropout & 3 & 5 & 4 & 3 & 5 & 4 \\
\hline
Adaptive Dropout & 5 & 3 & 5 & 4 & 3 & 4 \\
\hline
DropConnect & 4 & 2 & 3 & 5 & 4 & 3.6 \\
\hline
Uniform Dropout & 2 & 3 & 1 & 2 & 2 & 2 \\
\hline
Gaussian Dropout & 1 & 1 & 2 & 1 & 1 & 1.2 \\
\hline
\end{tabular*}
\end{table*}

During testing, the testing images are isotropically rescaled to a $256$ smallest image side,
denoted as Q. Then, the fully-connected layers are first converted to convolutional layers (
the first FC layer to a $7\times7$ conv. layer, the last two FC layers to $1\times1$
conv. layers). The resulting fully-convolutional net is then applied to the whole
(uncropped) image. The result is a class score map with the number of channels equal to
number of classes. Then, the class score map is spatially averaged (sum-pooled). And
the test set is also augmented by horizontal flipping of the images. Finally, the soft-max
class posteriors of the original and flipped images are averaged to obtain final scores for
the image as in \cite{VGG}.

Performance of all the dropout algorithms are shown in Table \ref{perf:ImageNet}.
This table shows that continuous dropout can improve the performance of conventional dropout algorithms even for very
large scale dataset. All the p-values are far less than 0.05, which indicates that Gaussian dropout achieves significantly
performance gain over other methods on this dataset.
%In summary, our proposed Gaussian dropout is a robust method that can boost the classification accuracy effectively.

%The large number of images and classes helps to train a model with high generalization ability
%and stability, therefore it is common for better model to achieve significant gain. But for such a complex model, covariance between
%neurons (features) is inevitable. In such cases, some intrinsic mechanism to
%ensure independent feature learning is important. Gaussian dropout is a very
%simple one.}

%\textcolor{red}{In conclusion, Gaussian dropout helps to improve the neural network
%models by reducing neuron covariance, especially when small models faced with
%underfitting or large models faced with inevitable feature redundancy.}

To summarize the overall performance of different dropout methods, we rank all five dropout methods according to their 
performance on each of the five datasets, as shown in Table \ref{perf:rank}. We can see that Gaussian dropout is ranked first on four datasets and ranked second
on one dataset.

%These results posits that continuous dropout statistically outperfoms other dropout
%methods on the image classification task under Wilcoxon signed rank test.

\section{Conclusion}
In this paper, we have introduced a new explanation for the dropout algorithm
from the perspective of the neural network properties in the human brain. The
activation rate of neurons in neural networks for different situations is random
and continuous. Inspired by this phenomenon, we extend the traditional binary
dropout to continuous dropout. Thorough theoretical analyses and extensive
experiments demonstrate that our continuous dropout has the advantage of reducing the
co-adaptation while maintaining variance, and continuous dropout is equivalent
to involving a regularizer that is able to prevent co-adaptation between
feature detectors.

In the future, we plan to further explore continuous dropout from the following two aspects.
First, although we have shown that continuous dropout penalizes the covariance between neurons,
the corresponding regularization term is not explicitly defined.
We will try to propose a more direct and interpretable way for the regularization term.
Second, dropout is naturally viewed as a mixture of different models. From this perspective of view,
we plan to derive an error bound for this way of mixture, leading to a more
solid theoretical analysis of continuous dropout.

%In the future, we plan to explore dropout from the following directions.
%First, Non dropout networks allow all the neurons
%in the same layer to cooperate for feature learning while dropout networks
%pushes all of them to do totally independent learning. A new approach which
%balances the aforementioned two mechanism could be interesting and more
%effective.
%Second, the regularization of covariance between neurons in continuous
%dropout is not explicit, we can propose a more direct and interpretable way.
%Third, dropout is naturally viewed as a mixture of models. From this point of view, we
%probably can derive an error bound for this way of mixture, leading to a more
%solid theoretical analysis of continuous dropout.

%\section*{Acknowledgment}

\ifCLASSOPTIONcaptionsoff
  \newpage
\fi

\bibliographystyle{IEEEtran}
\bibliography{mybib}


%\end{IEEEbiography}
%% insert where needed to balance the two columns on the last page with
%%\newpage


%\vfill


\end{document}